\documentclass[format=acmsmall, review=false, screen=true]{acmart}
\usepackage{times}
\usepackage{helvet}
\usepackage{courier}
\usepackage{amsmath,amsthm,amsfonts}

\usepackage{amssymb}

\usepackage{booktabs}

\usepackage{esvect}
\usepackage{mathrsfs}
\usepackage[font=small,labelfont=bf,labelsep=none]{caption}
\usepackage{amsmath}
\usepackage[noend]{algpseudocode}
\usepackage{bm}

\usepackage{graphicx}
\usepackage{mathrsfs}
\newcommand{\etal}{\emph{et~al.~}}

\acmJournal{JATS}
\acmVolume{32}
\acmNumber{14}
\acmArticle{21}
\acmMonth{2}

\begin{document}

    \title{Deep Reinforcement Learning for Unmanned Aerial Vehicle-Assisted Vehicular Networks}

    \author{Ming~Zhu}
    \email{zhumingpassional@gmail.com}
    \affiliation{
        \institution{Shenzhen Institutes of Advanced Technology, Chinese Academy of Sciences}
        \country{China}
    }

    \author{Xiao-Yang~Liu}
    \authornote{Both authors contributed equally to this research.}
    \email{xl2427@columbia.edu}
    \affiliation{
        \institution{Department of Electrical Engineering, Columbia University}
        \country{USA}
    }

    %\author{Sem Borst}
    %\email{sem.borst@outlook.com}
    %\affiliation{%
    %    \institution{Nokia Bell Labs}
    %    \country{USA}
    %}

    \author{Anwar Walid}
    \email{anwar.walid@nokia-bell-labs.com}
    \affiliation{%
        \institution{Nokia Bell Labs}
        \country{USA}
    }

    \begin{abstract}
        \noindent\rule[0.25\baselineskip]{\textwidth}{1pt}

        Unmanned aerial vehicles (UAVs) are envisioned to complement the 5G communication infrastructure in future smart cities. Hot spots easily appear in road intersections, where effective communication among vehicles is challenging. UAVs may serve as relays with the advantages of low price, easy deployment, line-of-sight links, and flexible mobility. In this paper, we study a UAV-assisted vehicular network where the UAV jointly adjusts its transmission control (power and channel) and 3D flight to maximize the total throughput. First, we formulate a Markov decision process (MDP) problem by modeling the mobility of the UAV/vehicles and the state transitions. Secondly, we solve the target problem using a deep reinforcement learning method, namely, the deep deterministic policy gradient (DDPG), and propose three solutions with different control objectives. Deep reinforcement learning methods obtain the optimal policy through the interactions with the environment without knowing the environment variables. Considering that environment variables in our problem are unknown and unmeasurable, we choose a deep reinforcement learning method to solve it. Moreover, considering the energy consumption of 3D flight, we extend the proposed solutions to maximize the total throughput per unit energy. To encourage or discourage the UAV's mobility according to its prediction, the DDPG framework is modified, where the UAV adjusts its learning rate automatically. Thirdly, in a simplified model with small state space and action space, we verify the optimality of proposed algorithms. Comparing with two baseline schemes, we demonstrate the effectiveness of proposed algorithms in a realistic model.

    \end{abstract}

    \begin{CCSXML}
<ccs2012>
   <concept>
       <concept_id>10003752.10010070.10010071.10010261</concept_id>
       <concept_desc>Theory of computation~Reinforcement learning</concept_desc>
       <concept_significance>500</concept_significance>
       </concept>
   <concept>
       <concept_id>10010405.10010481.10010485</concept_id>
       <concept_desc>Applied computing~Transportation</concept_desc>
       <concept_significance>500</concept_significance>
       </concept>
   <concept>
       <concept_id>10010405.10010481.10010484.10011817</concept_id>
       <concept_desc>Applied computing~Multi-criterion optimization and decision-making</concept_desc>
       <concept_significance>500</concept_significance>
       </concept>
 </ccs2012>
\end{CCSXML}
\ccsdesc[500]{Theory of computation~Reinforcement learning}
\ccsdesc[500]{Applied computing~Transportation}
\ccsdesc[500]{Applied computing~Multi-criterion optimization and decision-making}

    \keywords{
        Unmanned aerial vehicle, vehicular networks, smart cities, Markov decision process, deep reinforcement learning, power control, channel control.
    }

    \maketitle

\section{Introduction}\label{Sec:Introduction}

Intelligent transportation system \cite{chaqfeh2018DataDissemination} \cite{zhu2016PublicVehicle} is a key component of smart cities, which employs real-time data communication for traffic monitoring, path planning, entertainment and advertisement \cite{li2018CrowdTracking}. High speed vehicular networks \cite{cunha2016CommunicationVANET}  emerge as a key component of intelligent transportation systems that provide cooperative communications to improve data transmission performance.

With the increasing number of vehicles, the current communication infrastructure may not satisfy data transmission requirements, especially when hot spots (e.g., road intersections) appear during rush hours. Unmanned aerial vehicles (UAVs) or drones \cite{sedjelmaci2017UAV} can complement the 4G/5G communication infrastructure, including vehicle-to-vehicle (V2V) communications, and vehicle-to-infrastructure (V2I) communications. Qualcomm has received a certification of authorization allowing for UAV testing below 400 feet \cite{2018UAV_5G}; Huawei will cooperate with China Mobile to build the first cellular test network for regional logistics UAVs \cite{2018Drone_5G_Huawei}. Existing road side units (RSUs) and 5G stations can not adjust their 3D positions to obtain the best communication links since their positions are fixed. The energy consumption of UAVs can be very low compared with 5G stations and RSUs. The required energy in one day for 5G base station is 72 kWh (a macrocell) or 19.2 kWh (a small cell) \cite{ge2017energy}, which is much larger than UAVs \cite{2019UAV_DJI_MavicAir} and RSUs \cite{2020Huawei_RSU6201}.

\begin{figure}
  \centering
  \includegraphics[height=0.40\linewidth,width=0.65\linewidth]{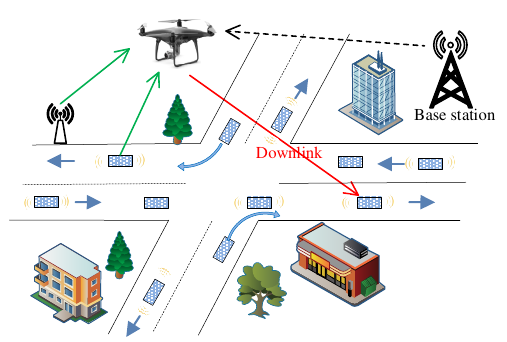}
  \caption{The scenario of a UAV-assisted vehicular network.}
  \label{Fig:Scenario}
\end{figure}

% The time of providing energy for UAVs is less than RSUs and 5G stations since UAVs are only used when current infrastructure cannot provide enough services, e.g., rush hours. In addition, there are more radio frequencies in air-to-ground links than ground-to-ground links (e.g., 5.9 GHZ for V2I \cite{chehri2019V2I}). The UAV can act as a mobile station to provide high communication performance for ground vehicles. In a UAV-assisted vehicular network, all communication links have two types of states: line-of-sight (LoS) and non-line-of-sight (NLoS). LoS/NLoS largely affects the communication performance since the channel power gain in NLoS links is much lower than LoS links. There are two types of actions for the UAV: three-dimensional (3D) movement, and communication control. First, the UAV adjusts its 3D position to obtain high LoS probability and better links. If the UAV is higher, the LoS probability will be larger. However, the channel power gain may become lower since the distance between the UAV and vehicles is longer. The UAV can also change two-dimensional (2D) positions so that it is nearer to the region with most vehicles to improve the total throughput. Secondly, the communication control is another control objective, which includes the power and channel control. The vehicles' movements are adjusted by the traffic lights. Therefore, the power and channel should be flexibly allocated according to the movement patterns of vehicles.

A UAV-assisted vehicular network has several advantages. First, the path loss will be much lower since the UAV can move nearer to vehicles compared with stationary base stations. Secondly, the UAV is flexible in adjusting the 3D position with the mobility of vehicles so that the communication performance \cite{alzenad2017UAV_Placement} is improved. Thirdly, the quality of UAV-to-vehicle links can be optimized by adjusting the UAV's 3D positions, and is generally better than that of terrestrial links \cite{giordani2018mmWave_5G}, since there will be more radio frequencies and the LoS can be flexibly  obtained. Existing road side units (RSUs) can not adjust their 3D positions to obtain the best communication links since they are fixed on road sides.

Maximizing the total throughput of UAV-to-vehicle links has several challenges. First, the communication channels vary with the UAV's 3D positions since various obstructions may lead to NLoS links, and the UAV's 3D position (especially height) affects the LoS/NLoS probability. Secondly, the joint adjustment of the UAV's 3D flight and transmission control (e.g., power control) cannot be solved directly using conventional optimization methods with unknown and unmeasurable environment variables. Thirdly, the channel conditions are hard to acquire, e.g., the path loss from the UAV to vehicles is closely related to the height/density of buildings and street width.

Existing works focuses on the UAV's two-dimensional (2D) path planning  \cite{garraffa2018UAV_Path}, or the communication control \cite{wang2018UAV_power}, or the joint control of them \cite{wu2018UAV_Trajectory} \cite{zeng2018UAV_Trajectory} \cite{zhang2018UAV_Trajectory}. Most of them relies on the accurate parameters, however, in 5G networks, it is not easy to measure the parameters accurately. In 5G networks, the strength of signals reduces significantly if the communication links are NLoS. The UAV can adjust its 3D position to obtain line-of-sight (NLoS) links. Generally, the UAV's height  largely affect the LoS/NLoS probability, therefore affects the communication performance. Some current  deep reinforcement learning (RL) based methods \cite{zhang2019RL_D2D_Relay} \cite{ning2019RL_vehicular}  \cite{shokry2020UAV_coverage}  can solve the UAV control or the communication control problem with unknown parameters. However, most of them do not consider the 3D movement of the UAVs, the movement patterns of vehicles under the control of the traffic lights, or the joint control of the above two types of actions. Besides, existing works assume the terminals are stationary, so that the movement patterns of terminals (e.g., vehicles) are not considered.

In this paper, we propose deep reinforcement learning based algorithms to maximize the total throughput of UAV-to-vehicle communications, which jointly adjusts the UAV's 3D flight and transmission control by learning through interacting with the environment. The UAV's 3D flight changes the 3D position so that better communication links can be obtained. The transmission control includes the power control and channel control aiming to improve the total throughput. The main contributions of this paper can be summarized as follows: 1) We formulate the problem as a Markov decision process (MDP) problem to maximize the total throughput with the constraints of total transmission power and total channel. Extracting the mobility patterns of UAV and vehicles is a challenge. The mobility patterns of vehicles in models should show that they are affected by the traffic light state. The mobility patterns of the UAV should consider the encoding/decoding of the horizontal flight and vertical flight. 2) We apply a deep reinforcement learning method, the deep deterministic policy gradient (DDPG), to solve the problem. DDPG is suitable to solve MDP problems with continuous states and actions. We propose three solutions with different control objectives to jointly adjust the UAV's 3D flight and transmission control. Then we extend the proposed solutions to maximize the total throughput per energy unit. To encourage or discourage the UAV's mobility, we modify the reward function and the DDPG framework. 3) We verify the optimality of proposed solutions using a simplified model with small state space and action space. Finally, we provide extensive simulation results to demonstrate the effectiveness of the proposed solutions compared with two baseline schemes. 

The remainder of the paper is organized as follows. Section \ref{Sec:RelatedWork} discusses related works. Section \ref{Sec:SystemModel} presents system models and problem formulation. Solutions are proposed in Section \ref{Sec:Solution}. Section \ref{Sec:PerformanceEvaluation} presents the performance evaluation. Section \ref{Sec:Conclusion} concludes this paper.

\section{2 Related Works} \label{Sec:RelatedWork}

The dynamic control for the UAV-assisted vehicular networks includes flight control and transmission control. Flight control mainly includes the planning of flight path, time, and direction. Yang \etal \cite{yang2018UAV_Path} proposed a joint genetic algorithm and ant colony optimization method to obtain the best UAV flight paths to collect sensory data in wireless sensor networks. To further minimize the UAVs' travel duration under certain constraints (e.g., energy limitations, fairness, and collision), Garraffa \etal \cite{garraffa2018UAV_Path} proposed a two-dimensional (2D) path planning method based on a column generation approach. Liu \etal \cite{liu2018UAV_RL} proposed a deep reinforcement learning approach to control a group of UAVs by optimizing the flying directions and distances to achieve the best communication coverage in the long run with limited energy consumption.

The transmission control of UAVs mainly concerns resource allocations, e.g., access selection, transmission power and bandwidth/channel allocation. Wang \etal \cite{wang2018UAV_power} presented a power allocation strategy for UAVs considering communications, caching, and energy transfer. In a UAV-assisted communication network, Yan \etal \cite{yan2018UAV_game} studied a UAV access selection and base station bandwidth allocation problem, where the interaction among UAVs and base stations was modeled as a Stackelberg game, and the uniqueness of a Nash equilibrium was obtained.

Joint control of both UAVs' flight and transmission has also be considered. Wu \etal \cite{wu2018UAV_Trajectory} considered maximizing the minimum achievable rates from a UAV to ground users by jointly optimizing the UAV's 2D trajectory and power allocation. Zeng \etal \cite{zeng2018UAV_Trajectory} proposed a convex optimization method to optimize the UAV's 2D trajectory to minimize its mission completion time while ensuring each ground terminal recovers the file with high probability when the UAV disseminates a common file to them. Zhang \etal \cite{zhang2018UAV_Trajectory} considered the UAV mission completion time minimization by optimizing its 2D trajectory with a constraint on the connectivity quality from base stations to the UAV. Fan \etal \cite{fan2018UAV_Placement_Resource} optimized the UAV's 3D flight and transmission control together; however, the 3D position optimization was converted to a 2D position optimization by the LoS link requirement. Most existing research works neglected adjusting UAVs' height to obtain better quality of links by avoiding various obstructions or NLoS links. This will reduce the service quality especially in cities especially with multiple viaducts, high buildings and trees.

Approximate dynamic programming (ADP) and stochastic dual dynamic programming (SDDP) can solve MDP problems with large space. Most ADP based approaches focus on path planning of UAVs or vehicles, e.g., how to avoid collision \cite{sunberg2016UAV_collision_avoidance} and coordinate a team of heterogeneous autonomous vehicles \cite{ferrari2011vehicle_ADP} \cite{zhu2018JointTransportationCharging}. SDDP \cite{downward2020SDDP} methods search the state space which may occur with large probability, so that the computing efficiency is improved. Both ADP and SDDP methods need all variables. However, measuring these variables in 5G networks requires a lot of labor and cost.

Deep reinforcement learning (DRL) based methods have been used in 5G vehicular networks \cite{zhu2019PathPlanning} to provide high quality-of-service (QoS) services. Challita \etal \cite{challita2018UAV_RL} proposed a deep reinforcement learning based method for a cellular UAV network by optimizing the 2D path and cell association to achieve a tradeoff between maximizing energy efficiency and minimizing both wireless latency and the interference on the path. A similar scheme is applied to provide intelligent traffic light control in \cite{Liu2018NIPS}. \cite{chen2020RL_mmWave_V2X} studies the temporal effects of dynamic blockage in vehicular networks and proposes a DRL method (DDPG) to overcome dynamic blockage. \cite{zhang2019RL_D2D_Relay} proposes a DRL method for a joint relay selection and power allocation problem in multihop 5G mmWave device to device transmissions. \cite{ning2019RL_vehicular} constructs an intelligent offloading framework for 5G-enabled vehicular networks by jointly utilizing licensed cellular spectrum and unlicensed channels. The deep double Q-learning network (DDQN) method is used to solve a subproblem, distributed cellular spectrum allocation. \cite{peng2020RL_Vehicular} studies joint allocation of the spectrum, computing, and storing resources in a multi-access edge computing based vehicular network. DDPG is used to solve this problem to satisfy the QoS requirements. \cite{chen2020RL_ResourceManagement} studies the age of information aware radio resource management for expected long-term performance optimization in a Manhattan grid vehicle-to-vehicle network. It decomposes the original MDP into a series of MDPs, and then uses long short-term memory and DRL to solve it. \cite{shokry2020UAV_coverage} introduces UAVs cell-free network for providing coverage to vehicles entering a highway that is not covered. It uses DRL (actor-critic method) to control UAVs to achieve efficient communication coverage. However, these methods do not consider the 3D movement of the agent (RSUs, or UAVs), the movement patterns of vehicle under the control of the traffic lights, or joint control of transmission power and channels. 

In addition, most existing works assumed that the ground terminals are stationary; whereas in reality, some ground terminals move with certain patterns, e.g., vehicles move under the control of traffic lights. This work studies a UAV-assisted vehicular network where the UAV's 3D flight and transmission control can be jointly adjusted, considering the mobility of vehicles in a road intersection.

\section{3 System Models and Problem Formulation}\label{Sec:SystemModel}

In this section, we first describe the traffic model and communication model, and then formulate the target problem as a Markov decision process. The variables in the communication model are listed in Table \ref{Tab:VariablesCommunicationModel} for easy reference.

\subsection{3.1 Road Intersection Model and Traffic Model} \label{Subsec:TrafficModel}

\begin{figure}
  \centering
  \includegraphics[height=0.40\linewidth,width=0.55\linewidth]{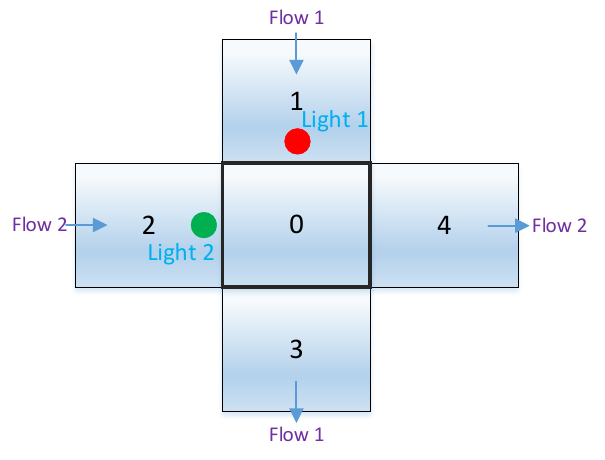}
  \caption{A one-way-two-flow road intersection.}
  \label{Fig:RoadIntersectionInModel}
\end{figure}

%For sake of transparency, we consider admittedly stylized traffic models that can capture the most essential features that reflect the dynamics of traffic flows in a road intersection. The realistic traffic models with multiple way cases in a road intersection is described in Subsection \ref{Subsec:RealisticTrafficModel}.

We start with a one-way-two-flow road intersection, as shown in Fig.~\ref{Fig:RoadIntersectionInModel}, while a much more complicated scenario in Fig.~\ref{Fig:RoadIntersection} will be described in Section \ref{Subsec:RealisticTrafficModel}. Five blocks are numbered as 0, 1, 2, 3, and 4, where block 0 is the intersection. We assume that each block contains at most one vehicle, indicated by binary variables $\bm{n} = (n^0, ..., n^4) \in \{ 0, 1 \}$. There are two traffic flows in Fig.~\ref{Fig:RoadIntersectionInModel},
\begin{itemize}
  \item{``Flow 1"}: $1 \rightarrow 0 \rightarrow 3$;
  \item{``Flow 2"}: $2 \rightarrow 0 \rightarrow 4$.
\end{itemize}

\begin{table}[tbp]
  \centering
  \caption{Variables in communication model}
  \label{Tab:VariablesCommunicationModel}
  \begin{tabular}{ l l }
  \hline
  $h^i_t, H^i_t$ & channel power gain and channel state from the UAV  \\
  & to a vehicle in block $i$ in time slot $t$. \\
  $\psi^i_t$ & Signal to interference and noise ratio (SINR) from the UAV \\
  & to a vehicle in block $i$ in time slot $t$. \\
  $d^i_t, D^i_t$ & horizontal distance and Euclidean distance between    \\
  & the UAV and a vehicle in block $i$. \\
  $P, C, b$ & total transmission power, total number of channels, and band-\\
  &width of each channel.\\
  $\rho^i_t, c^i_t$ & transmission power and number of channels allocated for the  \\
  & vehicle in block $i$ in time slot $t$. \\
  \hline
  \end{tabular}
\end{table}

\begin{figure}
  \centering
  \includegraphics[height=0.20\linewidth,width=0.65\linewidth]{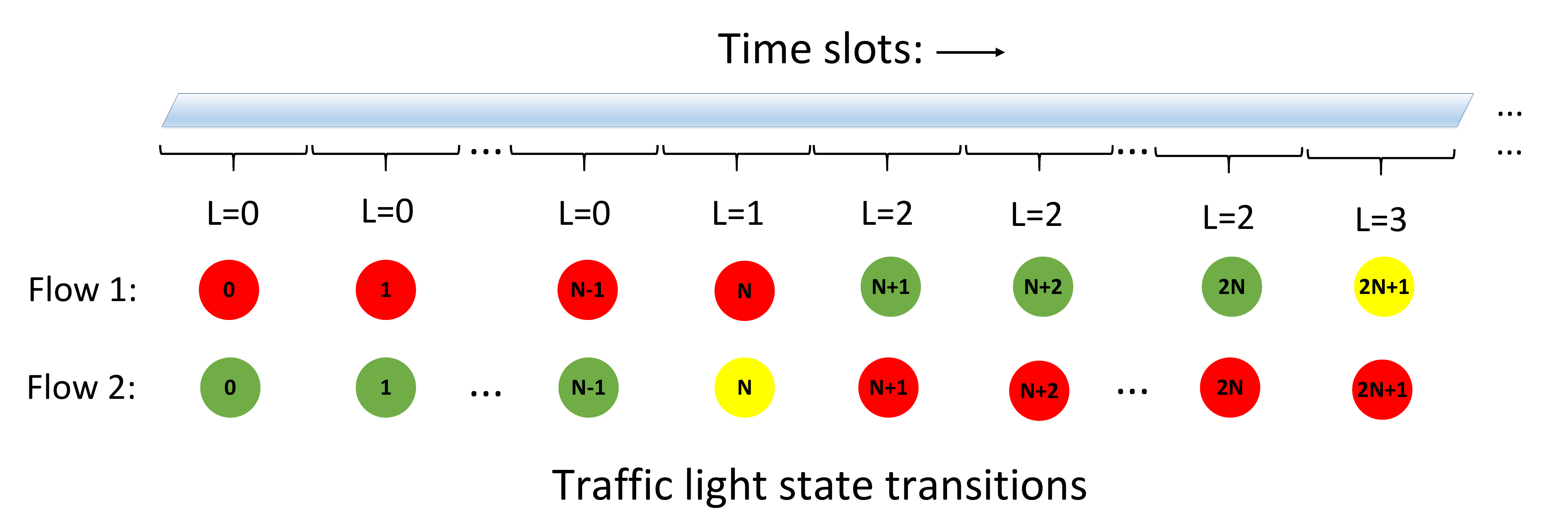}
  \caption{Traffic light states along time.}
  \label{Fig:TimeSlotTrafficLight}
\end{figure}

The traffic light $L$ has four configurations:
\begin{itemize}
\item{$L\!=\!0$}: red light for flow 1 and green light for flow 2;
\item{$L\!=\!1$}: red light for flow 1 and yellow light for flow 2;
\item{$L\!=\!2$}: green light for flow 1 and red light for flow 2;
\item{$L\!=\!3$}: yellow light for flow 1 and red light for flow 2.
\end{itemize}
Time is partitioned into slots with equal duration. The duration of a green or red light occupies $N$ time slots, and the duration of a yellow light occupies a time slot, which are shown in Fig.~\ref{Fig:TimeSlotTrafficLight}. We assume that each vehicle moves one block in a time slot if the traffic light is green.

\subsection{3.2 Communication Model} \label{Subsec:CommunicationModel}

We focus on the downlink communications (UAV-to-vehicle), since they are directly controlled by the UAV. There are two channel states of each UAV-to-vehicle link, line-of-sight (LoS) and non-line-of-sight (NLoS). Let $x$ and $z$ denote the block (horizontal position) and height of the UAV respectively, where $x \in \{ 0, 1, 2, 3, 4 \}$ corresponds to these five blocks in Fig.~\ref{Fig:RoadIntersectionInModel}, and $z$ is discretized to multiple values. We assume that the UAV stays above the five blocks since the UAV trends to get nearer to vehicles. Next, we describe the communication model, including the channel power gain, the signal to interference and noise ratio (SINR), and the total throughput.

First, the channel power gain between the UAV and a vehicle in block $i$ in time slot $t$ is $h^i_t$ with a channel state $H^i_t \in \{ \text{NLoS}, \text{LoS} \}$. $h^i$ is formulated as \cite{alzenad2017UAV_Placement} \cite{al2014OptimalAltitude}
\begin{eqnarray}
&& \hspace{-0.20in} h^i_t = \label{Eqn:ChannelPowerGain}
\begin{cases}
(D^i_t)^{-\beta_1},  ~~\,\, ~\text{if}~ H^i_t = \text{LoS},  \\
\beta_2 (D^i_t)^{-\beta_1}, ~\text{if}~ H^i_t = \text{NLoS},
\end{cases}
\end{eqnarray}
where $D^i_t$ is the Euclidean distance between the UAV and the vehicle in block $i$ in time slot $t$, $\beta_1$ is the path loss exponent, and $\beta_2$ is an additional attenuation factor caused by NLoS connections.

The probabilities of LoS and NLoS links between the UAV and a vehicle in block $i$ in time slot $t$ are \cite{mozaffari2016UAV_D2D}
\begin{eqnarray}
&& \hspace{-0.3in} p(H^i_t \! = \! \text{LoS}) \! = \! \frac{1}{1 + \alpha_1 \exp(-\alpha_2 (\frac{180}{\pi}\arctan \frac{z}{d^i_t} - \alpha_1))}, \label{Eqn:LOSProbability} \\
&& \hspace{-0.4in} p(H^i_t \! = \! \text{NLoS}) \! = \! 1 - p(H^i_t \! = \! \text{LoS}), ~i \in \{ 0, 1, 2, 3, 4 \}, \label{Eqn:NLOSProbability}
\end{eqnarray}
where $\alpha_1$ and $\alpha_2$ are system parameters depending on the environment (height/density of buildings, and street width, etc.). We assume that $\alpha_1$, $\alpha_2$, $\beta_1$, and $\beta_2$ have fixed values among all blocks in an intersection. $d^i_t$ is the horizontal distance in time slot $t$. The angle $\frac{180}{\pi} \arctan \frac{z}{d^i_t}$ is measured in ``degrees" with the range $0^{\circ} \sim 90^{\circ}$. Both $d^i_t$ and $z_t$ are discrete variables, therefore, $D^i_t = \sqrt{(d^i_t)^2 + z^2_t}$ is also a discrete variable.

Secondly, the SINR $\psi^i_t$ in time slot $t$ from the UAV to a vehicle in block $i$ is characterized as \cite{oehmann2015sinr}
\begin{eqnarray}
&& \psi^i_t = \frac{\rho^i_t h^i_t}{b c^i_t \sigma^2}, ~i \in \{ 0, 1, 2, 3, 4 \}, \label{Eqn:SINR}
\end{eqnarray}
where $b$ is the equal bandwidth of each channel, $\rho^i_t$ and $c^i_t$ are the allocated transmission power and number of channels for the vehicle in block $i$ in time slot $t$, respectively, and $\sigma^2$ is the additive white Gaussian noise (AWGN) power spectrum density, and $h^i$ is formulated by (\ref{Eqn:ChannelPowerGain}). We assume that the UAV employs orthogonal frequency division multiple access (OFDMA) \cite{gupta2016SubcarrierOFDM}; therefore, there is no interference among these channels.

Thirdly, the total throughput (reward) of UAV-to-vehicle links is formulated as \cite{ramezani2017throughput}
\begin{eqnarray}
&& \hspace{-0.45in} \sum_{i \in \{ 0, 1, 2, 3, 4 \}} \! b c^i_t \log (1 \!  + \!  \psi^i_t ) \! = \! \sum_{i \in \{ 0, 1, 2, 3, 4 \}} \! b c^i_t \log (1 \! + \! \frac{\rho^i_t h^i_t}{b c^i_t \sigma^2} ). \label{Eqn:TotalThroughput}
\end{eqnarray}

\subsection{3.3 MDP Formulation}\label{Sec:MDP}

The UAV aims to maximize the total throughput with the constraints of total transmission power and total channels:
\begin{eqnarray}
&& \sum_{i \in \{ 0, 1, 2, 3, 4 \}} \rho^i_t \leq P, ~\sum_{i \in \{ 0, 1, 2, 3, 4 \}} c^i_t \leq C,  \nonumber \\
&& 0 \leq \rho^i_t \leq \rho_{\text{max}}, ~~~~~~~~ 0 \leq c^i_t \leq c_{\text{max}}, ~i \in \{ 0, 1, 2, 3, 4 \}, \nonumber
\end{eqnarray}
where $P$ is the total transmission power, $C$ is the total number of channels, $\rho_{\text{max}}$ is the maximum power allocated to a vehicle, $c_{\text{max}}$ is the maximum number of channels allocated to a vehicle, $\rho^i_t$ is a discrete variable, and $c^i_t$ is a nonnegative integer variable. In MDP formulation, discrete transmission power is used for modeling and to help readers better understand the Markov property. Such a discrete formulation does not prevent a continuous solution. In fact, in the proposed algorithms, the transmission power is treated as continuous variable.

The UAV-assisted communication is modeled as a Markov decision process (MDP). The Markov property is that the future is independent of the past given the present. Let's discuss all state transition processes. 1) The next states of traffic lights depend on their current states, which is interpreted by Fig.~\ref{Fig:TimeSlotTrafficLight}. 2) The next 3D position of the UAV depends on its current 3D position and 3D flight action. 3) The next number of vehicles in each block depends on the vehicles' current positions, movement directions and the current states of traffic lights. 4) The channel states depend on the current 3D position of the UAV and the current positions of vehicles, which is reflected by \eqref{Eqn:LOSProbability} and \eqref{Eqn:NLOSProbability}. There are two types of stochastic processes. On one hand, from (\ref{Eqn:LOSProbability}) and (\ref{Eqn:NLOSProbability}), we know that the channel state of UAV-to-vehicle links follows a stochastic process. On the other hand, the arrival of vehicles follows a stochastic process under the states of the traffic lights.

Under the MDP framework, the state space $\mathcal{S}$, action space $\mathcal{A}$, reward $r$, policy $\pi$, and state transition probability $p(s_{t + 1}|s_t, a_t)$ of our problem can be defined. The MDP formulation can be extended to other network links with terminals moving with different speeds, e.g., the vehicle-to-RSU networks, the UAV networks, and the 5G station networks. 
\begin{itemize}
  \item{State} $\mathcal{S} = (L, x, z, \bm{n}, \bm{H})$, where $L$ is the traffic light state, $(x, z)$ is the UAV's 3D position with $x \in \{ 0, 1, 2, 3, 4 \}$ being the block and $z$ being the height, and $\bm{H} = (H^0, ..., H^4)$ is the channel state from the UAV to each block $i \in \{ 0, 1, 2, 3, 4 \}$ with $H^i \in \{ \text{NLoS}, \text{LoS} \}$. Let $z \in [z_{\text{min}}, z_{\text{max}}]$, where $z_{\text{min}}$ and $z_{\text{max}}$ are the UAV's minimum and maximum height, respectively. The block $x$ is the location projected from UAV's 3D position to the road. The channel state instead of the channel power gain is considered in the state space since the cost of testing the channel state is much lower than measuring the channel power gain. The received signal strength indicator (RSSI) is a type of radio interface. Based on RSSI, how to obtain the channel state is a classification problem. The channel state can be provided by machine learning \cite{huang2020machine}: ensemble learning methods (e.g., random forests, and gradient boosting), support vector machine, and deep neural networks. 
\end{itemize}
\textsl{}

\begin{figure}
  \centering
  \includegraphics[height=0.50\linewidth,width=0.53\linewidth]{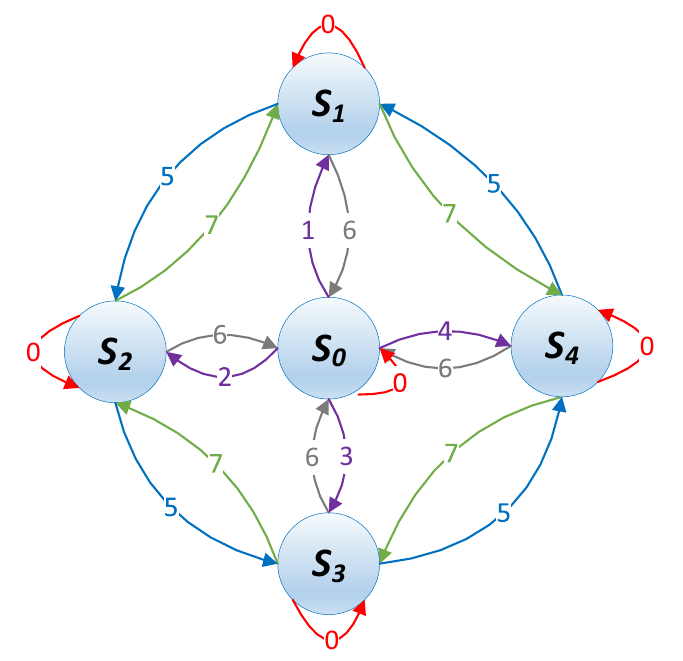}
  \caption{The position state transition diagram when the UAV's height is fixed.}
  \label{Fig:LocationStateTransitionDiagramInModel}
\end{figure}

\begin{itemize}
  \item{Action} $\mathcal{A} = (\bm{f}, \bm{\rho}, \bm{c})$ denotes the action set. $f^x$ denotes the horizontal flight, and $f^z$ denotes the vertical flight, both of which constitute the UAV's 3D flight $\bm{f} = (f^x, f^z)$. With respect to horizontal flight, we assume that the UAV can hover or flight to its adjacent block in a time slot, thus $f^x \in \{ 0, 1, ..., 7 \}$ in Fig.~\ref{Fig:LocationStateTransitionDiagramInModel}. With respect to vertical flight, we assume
      \begin{eqnarray}
      && f^z \in \{ -\nu, 0, \nu \}, \label{Eqn:UAVVerticalFlight}
      \end{eqnarray}
      which means that the UAV can fly downward $\nu$ meters, horizontally, and up $\nu$ meters in a time slot. The UAV's height changes as
      \begin{eqnarray}
      && z_{t + 1} = f^z_t + z_t. \label{Eqn:UAVHeightChange}
      \end{eqnarray}
      $\bm{\rho} = (\rho^0_t, ..., \rho^4_t)$ and $\bm{c} = (c^0_t, ..., c^4_t)$ are the transmission power and channel allocation actions for those five blocks, respectively. In the throughput optimization problem, the power and channel control actions are two basic actions, since they directly affect the communication performance in \eqref{Eqn:TotalThroughput}. The UAV's 3D flight is another action since it affects the LoS/NLoS probability of the links and the distance between the UAV and vehicles, therefore, it indirectly affects the communication performance. At the end of time slot $t$, the UAV moves to a new 3D position according to action $\bm{f}$, and over time slot $t$, the transmission power and number of channels are $\bm{\rho}$ and $\bm{c}$, respectively.
\end{itemize}
\begin{itemize}
  \item{Reward $r(s_t, a_t) = \sum_{i \in \{ 0, 1, 2, 3, 4 \}} b n^i_t c^i_t \log (1 + \frac{\rho^i_t h^i_t}{b c^i_t \sigma^2} )$} is the total throughput after a transition from state $s_t$ to $s_{t + 1}$ taking action $a_t$. Note that the total throughput over the $t$-th time slot is measured at the state $s_t = (L_t, x_t, z_t, \bm{n}_t, \bm{H}_t)$.
\end{itemize}
\begin{itemize}
  \item{Policy} $\pi$ is the strategy for the UAV, which maps states to a probability distribution over the actions $\pi: \mathcal{S} \rightarrow \mathcal{P}(\mathcal{A})$, where $\mathcal{P}(\cdot)$ denotes probability distribution. In time slot $t$, the UAV's state is $s_t = (L_t, x_t, z_t, \bm{n}_t, \bm{H}_t)$, and its policy $\pi_t$ outputs the probability distribution over the action $a_t$. We see that the policy indicates the action preference of the UAV.
\end{itemize}
\begin{itemize}
  \item{State transition probability $p(s_{t + 1}|s_t, a_t)$} formulated in (\ref{Eqn:TransitionProbability}) is the probability of the UAV entering the new state $s_{t + 1}$, after taking the action $a_t$ at the current state $s_t$. At the current state $s_t = (L_t, x_t, z_t, \bm{n}_t, \bm{H}_t)$, after taking the 3D flight and transmission control $a_t = (\bm{f}, \bm{\rho}, \bm{c})$, the UAV moves to the new 3D position $(x_{t + 1}, z_{t + 1})$, and the channel state changes to $\bm{H}_{t + 1}$, with the traffic light changes to $L_{t + 1}$ and the number of vehicles in each block changes to $\bm{n}_{t + 1}$.
\end{itemize}

The state transitions of the traffic light along time are shown in Fig.~\ref{Fig:TimeSlotTrafficLight}. The transition of the channel state for UAV-to-vehicle links is a stochastic process, which is reflected by (\ref{Eqn:LOSProbability}) and (\ref{Eqn:NLOSProbability}).

Next, we discuss the MDP in three aspects: the state transition probability, the state transitions of the number of vehicles in each block, and the UAV's 3D position. Note that the transmission power control and channel control do not affect the traffic light, the channel state, the number of vehicles, and the UAV's 3D position.

First, we discuss the state transition probability $p(s_{t + 1}|s_t, a_t)$ $=$ $p((L_{t + 1}, x_{t + 1}, z_{t + 1}, \bm{n}_{t + 1}, \bm{H}_{t + 1})$ $|(L_t, x_t, z_t, \bm{n}_t, \bm{H}_t)$, $(\bm{f}_t, \bm{\rho}_t, \bm{c}_t))$. The UAV's 3D fight only affects the UAV's 3D position state and the channel state, the traffic light state of the next time slot relies on the current traffic light state, and the number of vehicles in each block of the next time slot relies on the current number of vehicles and the traffic light state. Therefore, the state transition probability is
\begin{eqnarray}
&& p(s_{t + 1}|s_t, a_t) = p(x_{t + 1}, z_{t + 1}|x_t, z_t, \bm{f}_t) \nonumber \\
&& \hspace{0.93in} \times p(\bm{H}_{t + 1}|x_t, z_t, \bm{f}_t)  \times p(L_{t + 1}|L_t)  \nonumber \\
&& \hspace{0.93in} \times  p(\bm{n}_{t + 1}|L_t, \bm{n}_t), \label{Eqn:TransitionProbability}
\end{eqnarray}
where $p(x_{t + 1}, z_{t + 1}|x_t, z_t, \bm{f}_t)$ is easily obtained by the 3D position state transition based on the UAV's flight actions in Fig.~\ref{Fig:LocationStateTransitionDiagramInModel}, $p(\bm{H}_{t + 1}|x_t, z_t, \bm{f}_t)$ is easily obtained by (\ref{Eqn:LOSProbability}) and (\ref{Eqn:NLOSProbability}), $p(L_{t + 1}|L_t)$ is obtained by the traffic light state transition in Fig.~\ref{Fig:TimeSlotTrafficLight}, and $p(\bm{n}_{t + 1}|L_t, \bm{n}_t)$ is easily obtained by (\ref{Eqn:n0}) $\sim$ (\ref{Eqn:n1b}).

Secondly, we discuss the state transitions of the number of vehicles in each block. It is a stochastic process. The UAV's states and actions do not affect the number of vehicles of all blocks. Let $\lambda_1$ and $\lambda_2$ be the probabilities of the arrivals of new vehicles in flow 1 and 2, respectively.

The state transitions for the number of vehicles in block 0, 3, and 4 are
\begin{eqnarray}
&& \hspace{-0.098in} n^0_{t + 1} = \label{Eqn:n0}
\begin{cases}
n^2_t, ~\text{if}~ L_t = 0,    \\
n^1_t, ~\text{if}~ L_t = 2,    \\
0, ~\,\, ~\text{otherwise},
\end{cases}
\end{eqnarray}
\begin{eqnarray}
&& \hspace{0in} n^3_{t + 1} = \label{Eqn:n3}
\begin{cases}
n^0_t, \, ~\text{if}~ L_t = 2, 3,    \\
0, ~\, \, ~\text{otherwise},
\end{cases}
\end{eqnarray}
\begin{eqnarray}
&& \hspace{0in} n^4_{t + 1} = \label{Eqn:n4}
\begin{cases}
n^0_t,  \, ~\text{if}~ L_t = 0, 1,  \\
0, ~\, \, ~\text{otherwise}.
\end{cases}
\end{eqnarray}
The transition probability is 1 in (\ref{Eqn:n0}), (\ref{Eqn:n3}) and (\ref{Eqn:n4}) since the transitions are deterministic in block 0, 3, and 4. While the state transition probabilities for the number of vehicles in block 1 and 2 are nondeterministic, moreover, both of them are affected by their current number of vehicles and the traffic light. Taking block 1 when the traffic light state $L_t = 2$ as an example, the probability for the number of appeared vehicles is
\begin{eqnarray}
&& p(n^1_{t + 1} = 1|L_t = 2) = \lambda_1, \label{Eqn:n1a} \\
&& p(n^1_{t + 1} = 0|L_t = 2) = 1 - \lambda_1. \label{Eqn:n1b}
\end{eqnarray}
When $(n^1_t = 0, L_t \neq 2)$ and $(n^1_t = 1, L_t \neq 2)$, the probability for the number of vehicles will be obtained in a similar way.

Thirdly, we discuss the state transition of the UAV's 3D position. It includes horizontal position transitions and height transitions. The UAV's height transition is formulated in (\ref{Eqn:UAVHeightChange}). If the UAV's height is fixed, the horizontal position transitions are shown in Fig.~\ref{Fig:LocationStateTransitionDiagramInModel}, where $\{ S_i \}_{i \in \{ 0, 1, 2, 3, 4 \} }$ denotes the block (i.e, horizontal position) of the UAV: $0$ denotes staying in the current block; $\{ 1, 2, 3, 4 \}$ denotes a flight from block 0 to the other blocks (1, 2, 3, and 4); $5$ denotes an anticlockwise flight; $6$ denotes a flight from block 1, 2, 3, or 4 to block 0; $7$ denotes a clockwise flight.

\section{4 Proposed Solutions} \label{Sec:Solution}

In this section, we first describe the motivation, and then present an overview of Q-learning and the deep deterministic policy gradient (DDPG) algorithm, and then propose solutions with different control objectives, and finally present an extension of solutions that takes into account the energy consumption of 3D flight.

\subsection{4.1 Motivation}

The environment variables in this problem are unknown and unmeasurable. Deep reinforcement learning (RL) methods are suitable for the target problem for two reasons. First, DRL obtains the optimal policy through the interactions with the environment. Secondly, DRL does not need to know the environment variables. Neural networks are used to fit Q-value with the unknown and unmeasurable variables. For example, $\alpha_1$ and $\alpha_2$ are affected by the height/density of buildings, and the height and size of vehicles, etc. $\beta_1$ and $\beta_2$ are time dependent and are affected by the current environment such as weather \cite{agiwal2016_5G}. Although UAVs can detect the LoS/NloS links using equipped cameras, it is very challenging to detect them accurately for several reasons. First, the locations of recievers on vehicles should be labeled for detection. Secondly, it is hard to detect receivers accurately using computer vision technology since receivers are much small than vehicles. Thirdly, it requires automobile manufacturers to label the loactions of receivers, which may not be satisfied in several years. Therefore, it requires a large amount of labor to test these environment variables accurately.

It is hard to obtain the optimal strategies even all environment variables are known. Existing works \cite{wu2018UAV_throughput} \cite{zhang2018UAV_communication} obtain the near-optimal strategies in the 2D flight scenario when users are stationary, however, they are not capable of solving our target problem since the UAV adjusts its 3D position and vehicles move with their patterns under the control of traffic lights.

Q-learning cannot solve our problem because of several limitations. 1) Q-learning can only solve MDP problems with small state space and action space. However, the state space and action space of our problem are very large. 2) Q-learning cannot solve MDP problems with large state or action space since it may not converge within reasonable time. With respect to continuous state/action space, the state/action should be discretized if using Q-learning. The UAV's transmission power allocation actions are continuous. The discrete transmission power in Section III is simply to illustrate the MDP formulation. 3) Q-learning will converge slowly using too many computational resources \cite{sutton2018RL}, and this is not practical in our problem. Therefore, we adopt the DDPG algorithm to solve our problem.

\subsection{4.2 DDPG-based Algorithms}

The DDPG method \cite{lillicrap2016DDPG} uses deep neural networks to approximate both action policy $\pi$ and value function $Q(s, a)$. This method has two advantages: 1) it uses neural networks as approximators, essentially compressing the state and action space to much smaller latent parameter space, and 2) the gradient descent method can be used to update the network weights, which greatly speeds up the convergence and reduces the computational time. Therefore, the memory and computational resources are largely saved. In real systems, DDPG exploits the powerful skills introduced in AlphaGo zero \cite{silver2017Game} and Atari game playing \cite{mnih2013AtariDRL}, including experience replay buffer, actor-critic approach, soft update, and exploration noise.

\textbf{1) Experience replay buffer} $R_b$ stores transitions that will be used to update network parameters. At each time slot $t$, a transition $(s_t, a_t, r_t, s_{t + 1})$ is stored in $R_b$. After a certain number of time slots, each iteration samples a mini-batch of $M = |\Omega|$ transitions $\{ (s^j, a^j, r^j, s^j) \}_{j \in \Omega}$ to train neural networks, where $\Omega$ is a set of indices of sampled transitions from $R_b$. ``Experience replay buffer" has two advantages: 1) enabling the stochastic gradient decent method \cite{daniely2017SGD}; and 2) removing the correlations between consecutive transitions.

\textbf{2) Actor-critic approach}: the critic approximates the Q-value, and the actor approximates the action policy. The critic has two neural networks: the online Q-network $Q$ with parameter $\theta^Q$ and the target Q-network $Q'$ with parameter $\theta^{Q'}$. The actor has two neural networks: the online policy network $\mu$ with parameter $\theta^{\mu}$ and the target policy network $\mu'$ with parameter $\theta^{\mu'}$. The training of these four neural networks are discussed in the next subsection.

\textbf{3) Soft update} with a low learning rate $\tau \ll 1$ is introduced to improve the stability of learning. The soft updates of the target Q-network $Q'$ and the target policy network $\mu'$ are as follows
\begin{eqnarray}
&& \hspace{-0.2in} \theta^{Q'} \leftarrow \tau \theta^Q + (1 - \tau) \theta^{Q'} = \theta^{Q'} + \tau (\theta^Q - \theta^{Q'}), \label{Eqn:QUpdate} \\
&& \hspace{-0.2in} \theta^{\mu'}  \, \leftarrow \tau \theta^{\mu} + (1 - \tau) \theta^{\mu'} \,\, = \theta^{\mu'} + \tau (\theta^{\mu} - \theta^{\mu'}). \label{Eqn:MuUpdate}
\end{eqnarray}

\textbf{4) Exploration noise} is added to the actor's target policy to output a new action
\begin{eqnarray}
&& a_t = \mu'(s_t|\theta^{\mu'}) + \mathcal{N}_t. \label{Eqn:NewAction}
\end{eqnarray}
There is a tradeoff between exploration and exploitation, and the exploration is independent from the learning process. Adding exploration noise in (\ref{Eqn:NewAction}) ensures that the UAV has a certain probability of exploring new actions besides the one predicted by the current policy $\mu'(s_t|\theta^{\mu'})$, and avoids that the UAV is trapped in a local optimum.

\vspace{0.05in}
\vspace{0.05in}
\vspace{0.05in}
\begin{table*}[tbp]
  \normalsize
\begin{tabular}{lp{0.45\textwidth}}
  \toprule
  \textbf{Algorithm 1}: Environment simulation in one step (Fig.~\ref{Fig:LocationStateTransitionDiagramInModel} as an example) \\
  \toprule
  \,\,\,1:~Select action $a$ according to line 3 in Alg.~3; \\
  \,\,\,2:~Generate random variables: probabilities of LoS and NLoS according to \eqref{Eqn:LOSProbability} and \eqref {Eqn:NLOSProbability}, \\
  \,\,\,~~~respectively; and the probability for the number of appeared vehicles $\lambda$;\\
  \,\,\,3:~Executing action $a$, and determining the new state, including traffic light state according\\
  \,\,\,~~~to Fig.~\ref{Fig:TimeSlotTrafficLight}, the UAV's new 3D positions according to Fig.~\ref{Fig:LocationStateTransitionDiagramInModel} and \eqref{Eqn:UAVHeightChange}, and the number of vehicles\\
  \,\,\,~~~in blocks according to \eqref{Eqn:n0} $\sim$ \eqref{Eqn:n1b};
  \,\,\,4:~Calculate the reward $r$ using \eqref{Eqn:TotalThroughput}; \\
  \,\,\,5:~Update states, including traffic light state, UAV's 3D position and channel state, \\
  \,\,\,~~and the number of vehicles in blocks.\\
  \bottomrule
\end{tabular}
\end{table*}
\vspace{0.05in}
\vspace{0.05in}
\vspace{0.05in}

\vspace{0.05in}
\begin{table*}
\normalsize
\centering
\begin{tabular}[tbp]{lp{0.46\textwidth}}
  \toprule
  \textbf{Algorithm 2}: Channel allocation in time slot $t$ ~~~~~~~~~~~~~~~~~~~~~~~~~~~~~~~~~~~~~~~~~~~~~~~~~~~~~~~~~~~~~~~~~~~~~~~~~~~~~~~~~~~~~~~~\,\\
  \hline
  \,\,\,~\textbf{Input}:~the power allocation $\bm{\rho}$, the number of vehicles in all blocks $\bm{n}$, the maximum number \\
  \,\,\,~~~~~~~~~\,of channels allocated to a vehicle $c_{\text{max}}$, the total number of channels $C$.\\
  \,\,\,~\textbf{Output}:~the channel allocation $\bm{c}_t$ for all blocks. \\
  \,\,\,1:~Initialize the remaining total number of channels $C_r \leftarrow C$.\\
  \,\,\,2:~Calculate the average allocated power for each vehicle in all blocks $\bar{\bm{\rho}}_t$ by (\ref{Eqn:AveragePower}). \\
  \,\,\,3:~Sort $\bar{\bm{\rho}}_t$ by the descending order, and obtain a sequence of block indices $\bm{J}$. \\
  \,\,\,4:~\textbf{for} block $j \in \bm{J}$ \\
  \,\,\,5:~~~~~\hspace{0.1in}$c^j_t \leftarrow \min(C_r, n^j_t c_{\text{max}})$. \\
  \,\,\,7:~~~~~\hspace{0.1in}$C_r \leftarrow C_r - c^j_t$. \\
  \,\,\,8:~Return $\bm{c}_t$.\\
  \bottomrule
\end{tabular}
\end{table*}

\vspace{0.05in}
\begin{table*}
\normalsize
\centering
\begin{tabular}[tbp]{lp{0.46\textwidth}}
  \toprule
  \textbf{Algorithm 3}: DDPG-based algorithms: PowerControl, FlightControl, and JointControl  \\
  \hline
  \,\,\,~\textbf{Input}: the number of episodes $K$, the number of time slots $T$ in an episode, the mini-batch\\
  \,\,\,~size $M$, the learning rate $\tau$. \\
  \,\,\,1:~Initialize all states, including the traffic light state $L$, the UAV's 3D position $(x, z)$, \\
  \,\,\,~~~\hspace{0.1in}the number of vehicles $\bm{n}$ and the channel state $\bm{H}$ in all blocks. \\
  \,\,\,2:~Randomly initialize critic's online Q-network parameters $\theta^Q$ and actor's online policy network \\
  \,\,\,~~~\hspace{0.1in}parameters $\theta^{\mu}$, and initialize the critic's target Q-network parameters $\theta^{Q'} \leftarrow \theta^Q$ and actor's target \\
  \,\,\,~~~\hspace{0.1in}policy network parameters $\theta^{\mu'} \leftarrow \theta^{\mu}$. \\
  \,\,\,3:~Allocate an experience replay buffer $R_b$. \\
  \,\,\,4:~\textbf{for} episode $k = 1$ to $K$ \\
  \,\,\,5:~~~\hspace{0.1in}Initialize a random process (a standard normal distribution) $\mathcal{N}$ for the UAV's action exploration. \\
  \,\,\,6:~~~\hspace{0.1in}Observe the initial state $s_1$. \\
  \,\,\,7:~~~\hspace{0.1in}\textbf{for} $t = 1$ to T \\
  \,\,\,8:~~~~~\hspace{0.2in}Select the UAV's action $\bar{a}_t = \mu'(s_t|\theta^{\mu'}) + \mathcal{N}_t$ according to the policy of $\mu'$ and exploration noise $\mathcal{N}_t$. \\
  \,\,\,9:~~~~~\hspace{0.2in}\textbf{if} PowerControl \\
  10:~~~~~~~\hspace{0.3in}Combine the channel allocation in Alg.~1 and $\bar{a}_t$ as the UAV's action $a_t$ at a fixed 3D position. \\
  11:~~~~~\hspace{0.2in}\textbf{if} FlightControl \\
  12:~~~~~~~~~~\hspace{0.3in}Combine the equal transmission power, equal channel allocation and $\bar{a}_t$ (3D flight) as the \\
  ~~~~~~~~~~~~~\hspace{0.4in}UAV's action $a_t$. \\
  13:~~~~~\hspace{0.2in}\textbf{if} JointControl \\
  14:~~~~~~~~~~\hspace{0.3in}Combine the 3D flight action, the channel allocation in Alg.~1 and $\bar{a}_t$ as the UAV's action $a_t$. \\
  15:~~~~~\hspace{0.2in}Execute the UAV's action $a_t$, and receive reward $r_t$, and observe new state $s_{t+1}$ from the environment. \\
  16:~~~~~\hspace{0.2in}Store transition $(s_t, a_t, r_t, s_{t+1})$ in the UAV's experience replay buffer $R_b$. \\
  17:~~~~~\hspace{0.2in}Sample $R_b$ to obtain a random mini-batch of $M$ transitions $\{ (s^j_t, a^j_t, r^j_t, s^j_{t + 1}) \}_{j \in \Omega} \subseteq R_b$, where $\Omega$ is \\
  ~~~~~~~~\,\hspace{0.35in}a set of indices of sampled transitions with $|\Omega| = M$.\\
  18:~~~~~\hspace{0.2in}The critic's target Q-network $Q'$ calculates and outputs $y^j_t = r^j_t + \gamma Q'(s^j_{t + 1}, \mu'(s^j_{t + 1}| \theta^{\mu'})|\theta^{Q'})$ to  \\
  ~~~~~~~~\,\hspace{0.35in}the critic's online Q-network $Q$. \\
  19:~~~~~\hspace{0.2in}Update the critic's online Q-network $Q$ to make its Q-value fit $y^j_t$ by minimizing the loss function: \\
  ~~~~~~~~\,\hspace{0.35in}$\nabla_{\theta^Q} \text{Loss}_t(\theta^Q)= \nabla_{\theta^Q} [\frac{1}{M} \sum^{M}_{j = 1} (y^j_t - Q(s^j_t, a^j_t| \theta^Q))^2]$. \\
  20:~~~~~\hspace{0.2in}Update the actor's online policy network $\mu$ based on the input $\{ \nabla_a Q(s, a| \theta^Q)|_{s = s^j_t, a = \mu(s^j_t)} \}_{j \in \Omega}$ from \\
  ~~~~~~~~\,\hspace{0.35in}$Q$ using the policy gradient by the chain rule:\\
  ~~~~~~~~\,\hspace{0.35in}$\frac{1}{M} \sum_{j \in \Omega} \mathbb{E}_{s_t} [\nabla_{a} Q(s, a| \theta^Q)|_{s = s_t, a = \mu(s_t)} \nabla_{\theta^{\mu}} \mu(s|\theta^{\mu})|_{s = s_t} ]$.\\
  21:~~~~~\hspace{0.2in}Soft update the critic's target Q-network $Q'$ and actor's target policy network $\mu'$ to make the \\
  ~~~~~~~~\,\hspace{0.35in}evaluation of the UAV's actions and the UAV's policy more stable: \\
  ~~~~~~~~\,\hspace{0.35in}$\theta^{Q'}  \leftarrow \tau \theta^Q + (1 - \tau) \theta^{Q'}$, $\theta^{\mu'}  \leftarrow \tau \theta^{\mu} + (1 - \tau) \theta^{\mu'}$.\\
  \bottomrule
\end{tabular}
\end{table*}

\subsection{4.3 DDPG-based Solutions}

The UAV has two transmission controls, power and channel. We use the power allocation as the main control objective for two reasons. 1) Once the power allocation is determined, the channel allocation will be easily obtained in OFDMA. According to Theorem 4 of \cite{wang2015JointEnergyBandwidth}, in OFDMA, if all links have the equal weights just as our reward function (\ref{Eqn:TotalThroughput}), the transmitter should send messages to the receiver with the strongest channel in each time slot. In our problem, the strongest channel is not determined since the channel state (LoS or NLoS) is a random process. DDPG tends to allocate more power to the strongest channels with high probability, therefore, channel allocation will be easily obtained based on power allocation actions. 2) Power allocation is continuous, and DDPG is suitable to handle these actions. However, if we use DDPG for the channel allocation, the number of action variables will be very large and the convergence will be very slow, since the channel allocation is discrete and the number of channels is generally large (e.g., 200) especially in rush hours. We choose power control or flight as control objectives since controlling power and flight is more efficient than controlling channel. Moreover, the best channel allocation strategy can be obtained if the power is allocated in OFDMA. 

To allocate channels among blocks, we introduce a variable denoting the average allocated power of a vehicle in block $i$:
\begin{eqnarray}
&& \hspace{-0.098in} \bar{\rho}^i_t = \label{Eqn:AveragePower}
\begin{cases}
\frac{\rho^i_t}{n^i_t}, ~\text{if}~ n^i_t \neq 0,    \\
0, ~\,\, ~\text{otherwise}.
\end{cases}
\end{eqnarray}
The channel allocation algorithm is shown in Alg.~1, which is executed after obtaining the power allocation actions. As the above description, it achieves the best channel allocation in OFDMA if the power allocation is known \cite{wang2015JointEnergyBandwidth}. Line 1 is the initialization. Lines 2 $\sim$ 3 calculate and sort $\bar{\bm{\rho}}_t = \{ \bar{\rho}^i_t \}_{i \in \{ 0, 1, 2, 3, 4 \} }$. Line 5 assigns the maximum number of channels to the current possibly strongest channel, and line 6 updates the remaining total number of channels.

Based on the above analysis and Alg.~1, we propose three algorithms:
\begin{itemize}
\item{PowerControl:}
  the UAV adjusts the transmission power allocation using the actor network at a fixed 3D position, and the channels are allocated to vehicles by Alg.~1 in each time slot.
\end{itemize}
\begin{itemize}
\item{FlightControl:}
  the UAV adjusts its 3D flight using the actor network, and the transmission power and channel allocation are equally allocated to each vehicle in each time slot.
\end{itemize}
\begin{itemize}
\item{JointControl:}
  the UAV adjusts its 3D flight and the transmission power allocation using the actor network, and the channels are allocated to vehicles by Alg.~1 in each time slot.
\end{itemize}

The DDPG-based algorithms are given in Alg.~2. The algorithm has two parts: initializations, and the main process. Each episode finishes when the UAV executes a maximum number of steps, and the UAV's goal is to maximize the expected accumulated reward per episode. $s_1$ is the UAV's initial state. Generally, in each episode, $s_1$ is different so that the Q-value is trained well.

First, we describe the initializations in lines 1 $\sim$ 3. In line 1, all states are initialized: the traffic light $L$ is initialized as 0, the number of vehicles $\bm{n}$ in all blocks is 0, the UAV's block and height are randomized, and the channel state $H^i$ for each block $i$ is set as LoS or NLoS with the same probability. Note that the action space DDPG controls in PowerControl, FlightControl, and JointControl is different. Line 2 initializes the parameters of the critic and actor. Line 3 allocates an experience replay buffer $R_b$.

Secondly, we present the main process. Line 5 initializes a random process for action exploration. Line 6 receives an initial state $s_1$. Let $\bar{a}_t$ be the action DDPG controls, and $a_t$ be the UAV's all action. Line 8 selects an action according to $\bar{a}_t$ and an exploration noise $\mathcal{N}_t$. $\mathcal{N}_t$ is generated following a normal distribution. Lines 9 $\sim$ 10 combine the channel allocation actions in Alg.~1 and $\bar{a}_t$ as $a_t$ at a fixed 3D position in PowerControl. Lines 11 $\sim$ 12 combine the equal transmission power, equal channel allocation actions and $\bar{a}_t$ (3D flight) as $a_t$ in FlightControl. Lines 13 $\sim$ 14 combine the 3D flight action, the channel allocation actions in Alg.~1 and $\bar{a}_t$ as $a_t$ in JointControl. Line 15 executes the UAV's action $a_t$, and then the UAV receives a reward and all states are updated. Line 16 stores a transition into $R_b$. In line 17, a random mini-batch of transitions are sampled from $R_b$. Line 18 sets the value of $y^j$ for the critic's online Q-network. Lines 19 $\sim$ 21 update all network parameters.

How to obtain 3D flight and transmission power allocation in the action space in DDPG is an important issue. In fact, it is a mapping problem, i.e., how to encode and decode the two types of actions. The action space is separated to two parts, one for 3D flight (horizontal and the vertical flight) and the other for the transmission power allocation. The horizontal and the vertical flight actions are in a group, and the transmission power allocation actions are in the other group. Make sure that the locations of action variables match the physical meaning as much as possible. Take 8 horizontal flight actions in Fig.~2 as an example. We use 8 variables to denote the probability of them, and we will choose the one with the largest probability as the horizontal flight action.

The DDPG-based algorithms in Alg.~2 in essence are the approximated Q-learning method. The exploration noise in line 8 approximates the second case of (\ref{Eqn:EpsilonGreedy}) in Q-learning. Lines 18 $\sim$ 19 in Alg.~2 make $\left[ r_t + \gamma \max_{a_{t + 1}} Q(s_{t + 1}, a_{t + 1}) - Q(s_t, a_t) \right]$ converge. Line 20 of Alg.~2 approximates the first case of (\ref{Eqn:EpsilonGreedy}) in Q-learning, since both of them aim to obtain the policy of the maximum Q-value. In soft update of $Q'$ in line 21 of Alg.~2, $\tau$ and $\alpha$ are learning rates. Next, we discuss the training and test stages of proposed solutions.

%Fig.~\ref{Fig:DDPG_Framework} illustrates the data flow and parameter update process. 
\textbf{1)} In the training stage, we train the actor and the critic, and store the parameters of their neural networks. The training stage has two parts. First, $Q$ and $\mu$ are trained through a random mini-batch of transitions sampled from the experience replay buffer $R_b$. Secondly, $Q'$ and $\mu'$ are trained through soft update.

The training process is as follows. A mini-batch of $M$ transitions $\{ (s^j_t, a^j_t, r^j_t, s^j_{t + 1}) \}_{j \in \Omega}$ are sampled from $R_b$, where $\Omega$ is a set of indices of sampled transitions from $R_b$ with $|\Omega| = M$. Then two data flows are outputted from $R_b$: $\{ r^j_t, s^j_{t + 1} \}_{j \in \Omega} \rightarrow \mu'$, and $\{ s^j_t, a^j_t \}_{j \in \Omega} \rightarrow Q$. $\mu'$ outputs $\{ r^j_t, s^j_{t + 1}, \mu'(s^j_{t + 1}|\theta^{\mu'}) \}_{j \in \Omega}$ to $Q'$ to calculate $\{ y^j_t \}_{j \in \Omega}$. Then $Q$ calculates and outputs $\{ \nabla_{a} Q(s, a| \theta^Q)|_{s = s^j_t, a = \mu(s^j_t)} \}_{j \in \Omega}$ to $\mu$. $\mu$ updates its parameters by (\ref{Eqn:NablaQ}). Then two soft updates are executed for $Q'$ and $\mu'$ in (\ref{Eqn:QUpdate}) and (\ref{Eqn:MuUpdate}), respectively.

The data flow of the critic's target Q-network $Q'$ and online Q-network $Q$ are as follows. $Q'$ takes $\{ (r^j_t, s^j_{t + 1}, \mu'(s^j_{t + 1}|\theta^{\mu'})) \}_{j \in \Omega}$ as the input and outputs $\{ y^j_t \}_{j \in \Omega}$ to $Q$. $y^j_t$ is calculated by
\begin{eqnarray}
&&  y^j_t = r^j_t + \gamma Q'(s^j_{t + 1}, \mu'(s^j_{t + 1}| \theta^{\mu'})|\theta^{Q'}). \label{Eqn:y}
\end{eqnarray}
$Q$ takes $\left\{ \{ s^j_t, a^j_t \}_{j \in \Omega}, \right \}$ as the input and outputs $\{ \nabla_{a} Q(s, a| \theta^Q)|_{s = s^j_t, a = \mu(s^j_t)} \}_{j \in \Omega}$ to $\mu$ for updating parameters in (\ref{Eqn:NablaQ}), where $\{ s^j_t \}_{j \in \Omega}$ are sampled from $R_b$, and $\mu(s^j_t) = \arg\max_a Q(s^j_t, a)$.

The data flows of the actor's online policy network $\mu$ and target policy network $\mu'$ are as follows. After $Q$ outputs $\{ \nabla_{a} Q(s, a| \theta^Q)|_{s = s^j_t, a = \mu(s^j_t)} \}_{j \in \Omega}$ to $\mu$, $\mu$ updates its parameters by (\ref{Eqn:NablaQ}). $\mu'$ takes $\{ r^j_t, s^j_{t + 1} \}_{j \in \Omega}$ as the input and outputs $\{ r^j_t, s^j_{t + 1}, \mu'(s^j_{t + 1}|\theta^{\mu'}) \}_{j \in \Omega}$ to $Q'$ for calculating $\{ y^j_t \}_{j \in \Omega}$ in (\ref{Eqn:y}), where $\{ r^j_t, s^j_{t + 1} \}_{j \in \Omega}$ are sampled from $R_b$.

The updates of parameters of four neural networks ($Q$, $Q'$, $\mu$, and $\mu'$) are as follows. The online Q-network $Q$ updates its parameters by minimizing the $L_2$-norm loss function $\text{Loss}_t(\theta^Q)$ to make its Q-value fit $y^j_t$:
\begin{eqnarray}
&& \hspace{-0.4in} \nabla_{\theta^Q} \text{Loss}_t(\theta^Q)= \nabla_{\theta^Q} [\frac{1}{M} \sum^{M}_{j = 1} (y^j_t - Q(s^j_t, a^j_t| \theta^Q))^2]. \label{Eqn:Loss}
\end{eqnarray}
The target Q-network $Q'$ updates its parameters $\theta^{Q'}$ by (\ref{Eqn:QUpdate}). The online policy network $\mu$ updates its parameters following the chain rule with respect to $\theta^{\mu}$:
\begin{eqnarray}
&& \hspace{-0.46in} \mathbb{E}_{s_t} [\nabla_{\theta^{\mu}} Q(s, a| \theta^Q)|_{s = s_t, a = \mu(s_t|\theta^{\mu})}] \nonumber \\
&& \hspace{-0.29in} = \mathbb{E}_{s_t} [\nabla_{a} Q(s, a| \theta^Q)|_{s = s_t, a = \mu(s_t)} \nabla_{\theta^{\mu}} \mu(s|\theta^{\mu})|_{s = s_t} ].
\label{Eqn:NablaQ}
\end{eqnarray}
The target policy network $\mu'$ updates its parameters $\theta^{\mu'}$ by (\ref{Eqn:MuUpdate}).

In each time slot $t$, the current state $s_t$ from the environment is delivered to $\mu'$, and $\mu'$ calculates the UAV's target policy $\mu'(s_t)|\theta^{\mu'}$. Finally, an exploration noise $\mathcal{N}$ is added to $\mu'(s_t|\theta^{\mu'})$ to get the UAV's action in (\ref{Eqn:NewAction}).

\textbf{2)} In the test stage, we restore the neural network of the actor's target policy network $\mu'$ based on the stored parameters. This way, there is no need to store transitions to the experience replay buffer $R_b$. Given the current state $s_t$, we use $\mu'$ to obtain the UAV's optimal action $\mu'(s_t|\theta^{\mu'})$. Note that there is no noise added to $\mu'(s_t|\theta^{\mu'})$, since all neural networks have been trained and the UAV has got the optimal action through $\mu'$. Finally, the UAV executes the action $\mu'(s_t|\theta^{\mu'})$.

\subsection{4.4 Extension on Energy Consumption of 3D Flight}

The UAV's energy is used in two parts, communication and 3D flight. The above proposed solutions in Alg.~2 do not consider the energy consumption of 3D flight. In this subsection, we discuss how to incorporate the energy consumption of 3D flight into Alg.~2. To encourage or discourage the UAV's 3D flight actions in different directions with different amount of energy consumption, we modify the reward function and the DDPG framework.

The UAV aims to maximize the total throughput per energy unit since the UAV's battery has limited capacity. For example, the UAV DJI Mavic Air \cite{2019UAV_DJI_MavicAir} with full energy can only fly 21 minutes. Given that the UAV's energy consumption of 3D flight is much larger than that of communication, we only use the former part as the total energy consumption. Thus, the reward function (\ref{Eqn:TotalThroughput}) is modified as follows
\begin{eqnarray}
&& \hspace{-0.45in} \bar{r}(s_t, a_t) = \frac{1}{e(a_t)} \sum_{i \in \{ 0, 1, 2, 3, 4 \}} b n^i_t c^i_t \log (1 + \frac{\rho^i_t h^i_t}{b c^i_t \sigma^2} ), \label{Eqn:RewardExtension}
\end{eqnarray}
where $e(a_t)$ is the energy consumption of taking action $a_t$ in time slot $t$. $e(a_t)$ is calculated according to the UAV's energy consumption model for 3D flight. Alternatively, $e(a_t)$ can be measured easily by the following way. The UAV has three vertical flight actions per time slot just as in (\ref{Eqn:UAVVerticalFlight}). If the UAV keeps moving downward, horizontally, or upward until the energy for 3D flight is used up, the flight time is assumed to be $\phi_d$, $\phi_h$, and $\phi_u$ seconds, respectively. If the duration of a time slot is set to $\kappa$ seconds, so the UAV can fly $\frac{\phi_d}{\kappa}$, $\frac{\phi_h}{\kappa}$, and $\frac{\phi_u}{\kappa}$ time slots, respectively. Therefore, the formulation of $e(a_t)$ is given by
\begin{eqnarray}
&& \hspace{-0.098in} e(a_t) = \label{Eqn:EnergyConsumption}
\begin{cases}
\frac{\kappa}{\phi_d} E_{\text{full}}, ~\text{if~moving downward $\nu$ meters},    \\
\frac{\kappa}{\phi_h} E_{\text{full}}, ~\text{if moving horizontally},    \\
\frac{\kappa}{\phi_u} E_{\text{full}}, ~\text{if moving upward $\nu$ meters},
\end{cases}
\end{eqnarray}
where $E_{\text{full}}$ is the total energy if the UAV's battery is full.

Let $\delta(t)$ be a prediction error as follows
\begin{eqnarray}
&& \delta(t) = \bar{r}(s_t, a_t) - Q(s_t, a_t), \label{Eqn:Delta}
\end{eqnarray}
where $\delta(t)$ denotes the difference between the actual reward $\bar{r}(s_t, a_t)$ and the expected return $Q(s_t, a_t)$. The prediction error is different from that in machine learning such as ensemble learning. In ensemble learning, the prediction error is the difference between predicted value and actual value. The learning rates here are adjusted following the principle: when the prediction error is non-negative, a higher learning rate will be used, and a lower learning rate will be used otherwise. To make the UAV learn from the prediction error $\delta(t)$, not the difference between the new Q-value and old Q-value in (\ref{Eqn:UpdateQInQLearning}), the Q-value is updated by the following rule
\begin{eqnarray}
&& \hspace{-0.2in} Q(s_t, a_t) \leftarrow Q(s_t, a_t) + \alpha \delta(t)  \Leftrightarrow \nonumber \\
&& \hspace{-0.2in} Q(s_t, a_t) \leftarrow Q(s_t, a_t) + \alpha (\bar{r}(s_t, a_t) - Q(s_t, a_t)),
\label{Eqn:UpdateQInExtension}
\end{eqnarray}
where $\alpha$ is a learning rate similar to (\ref{Eqn:UpdateQInQLearning}).

We introduce $\alpha^+$ and $\alpha^-$ to represent the learning rate when $\delta(t) \geq 0$ and $\delta(t) < 0$, respectively. Therefore, the UAV can choose to be active or inactive by properly setting the values of $\alpha^+$ and $\alpha^-$. The update of Q-value in Q-learning is modified as follows, inspired by \cite{lefebvre2017optimisticRL}
\begin{eqnarray}
&& \hspace{-0.4in} Q(s_t, a_t) \leftarrow Q(s_t, a_t) +
\begin{cases}
\alpha^+ \delta(t),  ~\text{if}~\delta(t) \geq 0,  \\
\alpha^- \delta(t),  ~\text{if}~\delta(t) < 0.
\end{cases}
\end{eqnarray}

We define the prediction error $\delta(t)$ as the difference between the actual reward and the output of the critic's online Q-network $Q$:
\begin{eqnarray}
&& \delta(t) = \bar{r}(s_t, a_t) - Q(s_t, a_t|\theta^Q). \label{Eqn:Delta_DDPG}
\end{eqnarray}
We use $\tau^+$ and $\tau^-$ to denote the weights when $\delta(t) \geq 0$ and $\delta(t) < 0$, respectively. The update of the critic's target Q-network $Q'$ is
\begin{eqnarray}
&& \hspace{-0.4in} \theta^{Q'}  \leftarrow \label{Eqn:QUpdateExtension}
\begin{cases}
\tau^+ \theta^Q + (1 - \tau^+) \theta^{Q'} ,  ~\text{if}~\delta(t) \geq 0,  \\
\tau^- \theta^Q + (1 - \tau^-) \theta^{Q'} ,  ~\text{if}~\delta(t) < 0.
\end{cases}
\end{eqnarray}
The update of the actor's target policy network $\mu'$ is
\begin{eqnarray}
&& \hspace{-0.4in} \theta^{\mu'}  \leftarrow \label{Eqn:MuUpdateExtension}
\begin{cases}
\tau^+ \theta^{\mu} + (1 - \tau^+) \theta^{\mu'} ,  ~\text{if}~\delta(t) \geq 0,  \\
\tau^- \theta^{\mu} + (1 - \tau^-) \theta^{\mu'} ,  ~\text{if}~\delta(t) < 0.
\end{cases}
\end{eqnarray}

If $\tau^+ > \tau^-$, the UAV is active and prefers to move. If $\tau^+ < \tau^-$, the UAV is inactive and prefers to stay. If $\tau^+ = \tau^-$, the UAV is neither active nor inactive. To approximate the Q-value, we introduce $\bar{y}^j_t$ similar to (\ref{Eqn:y}) and then make the critic's online Q-network $Q$ to fit it. We optimize  the loss function
\begin{eqnarray}
&& \hspace{-0.4in} \nabla_{\theta^Q} \text{Loss}_t(\theta^Q)= \nabla_{\theta^Q} [\frac{1}{M} \sum^{M}_{j = 1} (\bar{y}^j_t - Q(s^j_t, a^j_t| \theta^Q))^2 ],
\end{eqnarray}
where $\bar{y}^j_t = \bar{r}^j_t$.

We modify the MDP, DDPG framework, and DDPG-based algorithms by considering the energy consumption of 3D flight:
\begin{itemize}
  \item{} The MDP is modified as follows. The state space $\mathcal{S} = (L, x, z, n, H, E)$, where $E$ is the energy in the UAV's battery. The energy changes as follows
  \begin{eqnarray}
  && E_{t + 1} = \max \{ E_t - e(a_t), 0 \}. \label{Eqn:EnergyChange}
  \end{eqnarray}
The other parts of MDP formulation and state transitions are the same as in Section \ref{Sec:MDP}.
\end{itemize}
\begin{itemize}
  \item{} To encourage or discourage the UAV's mobility by the predictions, we modify the DDPG framework so that the UAV adjusts its movement mode by changing the learning rate according to the prediction error $\delta(t)$. There are three modifications in the DDPG framework: a) The critic's target Q-network $Q'$ feeds $\bar{y}^j = \bar{r}^j$ to the critic's online Q-network $Q$ instead of $y^j$ in (\ref{Eqn:y}). b) The update of the critic's target Q-network $Q'$ is (\ref{Eqn:QUpdateExtension}) instead of (\ref{Eqn:QUpdate}). c) The update of the actor's target policy network $\mu'$ is (\ref{Eqn:MuUpdateExtension}) instead of (\ref{Eqn:MuUpdate}).
\end{itemize}
\begin{itemize}
  \item{} The DDPG-based algorithms are modified from Alg.~2. Initialize the energy state of the UAV as full in the start of each episode. In each time step of an episode, the energy state is updated by (\ref{Eqn:EnergyChange}), and this episode terminates if the energy state $E_t \leq 0$. The reward function is replaced by (\ref{Eqn:RewardExtension}).
\end{itemize}

\section{5 Performance Evaluation} \label{Sec:PerformanceEvaluation}

In this section, we first verify the optimality of DDPG-based algorithms in a simple road intersection. Then, we consider a complex road intersection. 

%We build a more realistic traffic model, describe baseline schemes and show simulation results.

Our experiments are executed on a server with Linux OS, $200$ GB memory, two Intel(R) Xeon(R) Gold 5118 CPUs@2.30 GHz, a Tesla V100-PCIE GPU.

\subsection{5.1 Optimality Verification} \label{Subsec:OptimalityVerification}

The implementation of Alg.~3 includes two parts: building the environment (including traffic and communication models) for our scenarios, and using DDPG in TensorFlow \cite{abadi2016tensorflow}. In the simulations, we apply a 4-layer fully-connected neural network for both the critic and actor with 100 neurons for the first 2 layers, and the rest two layers with 200 and 50 neurons, respectively. This is because the state and action space is large. We use the ``leaky ReLU" as the activation function, since it allows a small positive gradient when the unit is not active.

\textbf{Methodology}: DRL algorithms are black-box methods. The optimal solution can be obtained using conventional MDP methods in a small state-action space, e.g., policy iteration in MDP Toolbox \cite{Python_MDP}. Therefore, we choose a simple scenario in Fig.~\ref{Fig:RoadIntersectionInModel} to verify the optimality. If the results of DRL algorithms match the optimal solution of conventional MDP methods. We can conclude that the proposed DRL algorithms achieve the optimality. 

We make several assumptions for the scenario in Fig.~\ref{Fig:RoadIntersectionInModel} to keep the state-action space small for verification purpose. We assume the channel states of all communication links are LoS, and the UAV's height is fixed as 150 meters, so that the UAV can only adjust its horizontal flight control and transmission control. The traffic light state is assumed to have two values (red or green).

%We provide the optimality verification of DDPG-based algorithms in Alg.~3 in a one-way-two-flow road intersection in Fig.~\ref{Fig:RoadIntersectionInModel}. The reasons are as follows: (i) the MDP problem in such a simple scenario is explicitly defined and the theoretically optimal policy can be obtained using the Python MDP Toolbox \cite{Python_MDP}; and (ii) this optimality verification process also serves a good code debugging process before we apply the DDPG algorithm in TensorFlow \cite{abadi2016tensorflow} to the more realistic road intersection scenario in Fig.~\ref{Fig:RoadIntersection}.

\textbf{Experimental settings}: In Fig.~\ref{Fig:RoadIntersectionInModel}, we describe the environmental parameters used in Alg.~1. The values of parameters in simulating environments are summarized in Table~\ref{Tab:ValuesSimulating}. These values are unknown to the UAV. The UAV obtains the optimal policy through a number of transitions (the current state, action, the reward, and the next state) instead of these parameters. There are two types of parameters, communication parameters and UAV/vehicle parameters.

First, we describe communication parameters. $\alpha_1$ and $\alpha_2$ are set to 9.6 and 0.28, which are common values in urban areas \cite{mozaffari2015UAV_LOS}. $\beta_1$ is 3, and $\beta_2$ is 0.01, which are widely used in path loss modeling. The duration of a time slot is set to 6 seconds, and the number of occupied red or green traffic light $N$ is 10, i.e., 60 seconds constitute a red/green duration, which is commonly seen in cities and can ensure that the vehicles in blocks can get the next block in a time slot. The white power spectral density $\sigma^2$ is set to -130 dBm/Hz. 

Secondly, we describe UAV/vehicle parameters. We assume the arrival of vehicles in block 1 and 2 follows a binomial distribution with the same parameter $\lambda$ in the range $0.1 \sim 0.7$. The length of a road block $\widehat{d}$ is set to 3 meters. The blocks' distance is easily calculated as follows: $D(1, 0) = \widehat{d}$, and $D(1, 3) = 2 \widehat{d}$, where $D(i, j)$ is the Euclidean distance from block $i$ to block $j$. $\nu$ is set to 5 meters, which is common among UAVs. The energy consumption setups for the UAV follow DJI Mavic Air \cite{2019UAV_DJI_MavicAir}: $\phi_d$, $\phi_h$, and $\phi_u$ are set to 27 * 60, 21 * 60, and 17 * 60 seconds. The duration of a time slot $\kappa$ is set to 6 seconds.

The values of configures are summarized in Table~\ref{Tab:ValuesConfigures}. There are two types of configures: DDPG algorithm configures, and communication configures. First, we describe the DDPG algorithm configures. The number of episodes is 256, and the number of time slots in an episode is 256, so the number of total time slots is 65,536. The experience replay buffer capacity is 10,000, and the learning rate of target networks $\tau$ is 0.001. The mini-batch size $M$ is $512$. The training data set is full in the $10,000^{th}$ time slot, and is updated in each of the following 256 $\times$ 256 - 10,000 = 55,536 time slots. The test data set is real-time among all the 256 $\times$ 256 = 65,536 time slots. The discount factor $\gamma$ is 0.9. 

Secondly, we describe communication configures. The total UAV transmission power $P$ is set to $6$ W in consideration of the limited communication ability. The total number of channels $C$ is 10. The bandwidth of each channel $b$ is 100 KHz. Therefore, the total bandwidth of all channels is 1 MHz. The maximum power allocated to a vehicle $\rho_{\text{max}}$ is 3 W, and the maximum number of channels allocated to a vehicle $c_{\text{max}}$ is 5. We assume that the power control for each vehicle has 4 discrete values (0, 1, 2, 3).

%The configure of neural networks in proposed solutions is based on the configure of the DDPG action space. A neural network consists of an input layer, fully-connected layers, and an output layer. The number of fully-connected layers in actor is set to 4.

\begin{table}[tbp] 
  \centering
  \caption{Values of parameters in simulating environments}
  \label{Tab:ValuesSimulating}
  \begin{tabular}{ c c c c c c}
  \hline
  $\alpha_1$ & $\alpha_2$ & $\beta_1$ & $\beta_2$ & $\sigma^2$ & $\nu$  \\
  0.28 & 3 & 0.01 & -130 dBm/Hz & 3 & 5    \\
  \hline
  $\lambda$  & $g^s_i$  & $g^l_i$  & $g^r_i$ & $\phi_u$ & $\kappa$ \\
  0.1 $\sim$ 0.7  &  0.4 & 0.3  & 0.3 & 17 * 60 & 6\\
  \hline
  $\widehat{d}$ & $\phi_d$ & $\phi_h$ & $N$\\
  9.6  & 27 * 60 & 21 * 60 & 10 \\
  \hline
  \end{tabular}
\end{table}

\begin{table}[tbp] 
  \centering
  \caption{Values of configures}
  \label{Tab:ValuesConfigures}
  \begin{tabular}{ c c c c c c }
  \hline
  $P$ &  $C$  &  $\gamma$ & $z_{\text{min}}$ & $z_{\text{max}}$ \\
  1 $\sim$ 6 & 10 & 0.9 & 10 & 200  \\
  \hline
  $M$  &  $b$ & $\tau$ & $\rho_{\text{max}}$ &  $c_{\text{max}}$ \\
  512 & 100 KHz & 0.001 & 3 W & 5\\
  \hline
  \end{tabular}
\end{table}

\begin{figure*}
  \centering
  \begin{minipage}{0.46\textwidth}
    \includegraphics[height=0.98\linewidth,width=0.98\linewidth]{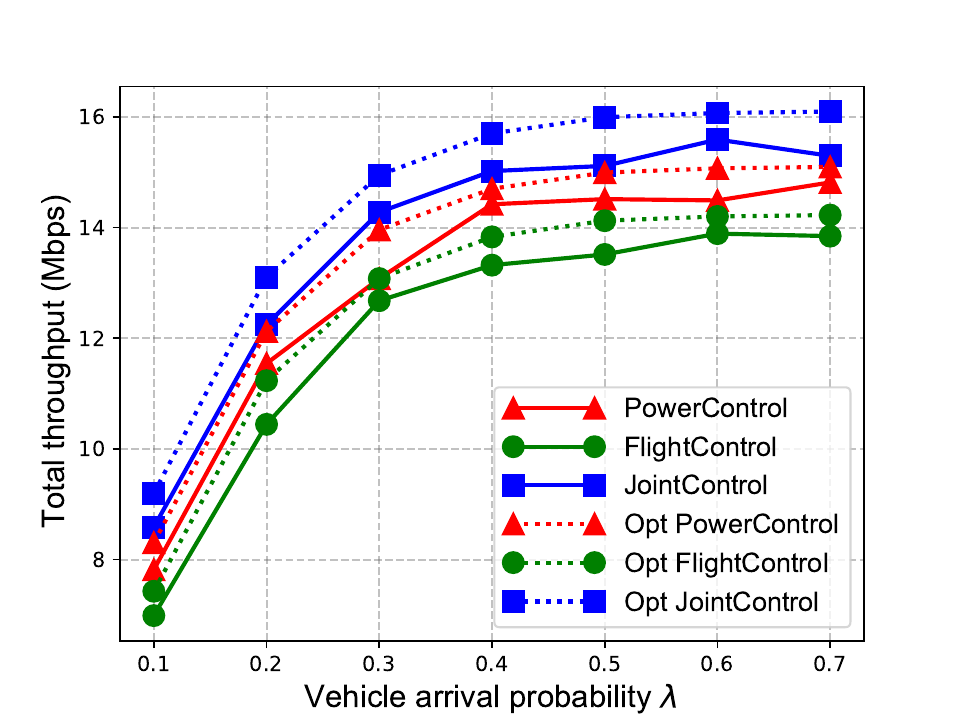}
    \caption{Total throughput vs. vehicle arrival probability $\lambda$ in optimality verification.}
    \label{Fig:ThroughputWithLambdaDDPGAndOptimal}
  \end{minipage}
  \hspace{0.2in}
  \begin{minipage}{0.46\textwidth}
    \includegraphics[height=0.98\linewidth,width=0.98\linewidth]{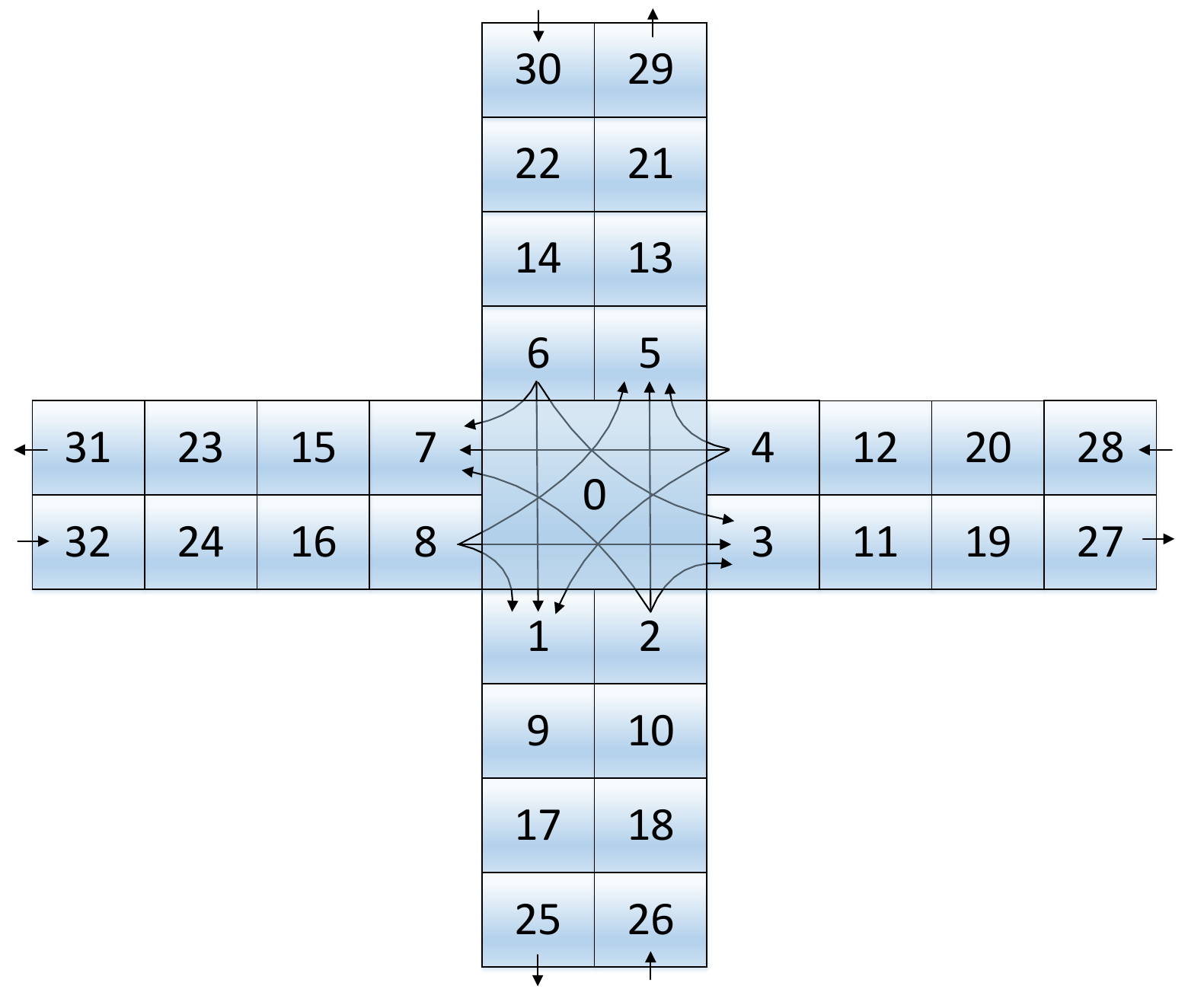}
    \caption{Realistic road intersection model.}
    \label{Fig:RoadIntersection}
  \end{minipage}
\end{figure*}

\textbf{Results}: The total throughput obtained by the policy iteration algorithm and DDPG-based algorithms are shown as dashed lines and solid lines in Fig.~\ref{Fig:ThroughputWithLambdaDDPGAndOptimal}. Therefore, DDPG-based algorithms achieve near optimal policies. We see that, the total throughput in JointControl is the largest, which is much higher than PowerControl and FlightControl. This is in consistent with our believes that the JointControl of power and flight allocation will be better than the control of either of both. The performance of PowerControl is better than FlightControl. The throughput increases with the increasing of vehicle arrival probability $\lambda$ in all algorithms, and it saturates when $\lambda \geq 0.6$ due to traffic congestion. 

The result of proposed algorithms matches that of the optimal policy. Therefore, we get a conclusion that the DDPG-based algorithms achieve the optimality in the simple road in Fig.~\ref{Fig:RoadIntersectionInModel}. 

% Theoretically, it is well-known that DRL algorithms (including DDPG algorithms) solve MDP problems and achieve the optimal results with much less memory and computational resources. 

\subsection{5.2 More Realistic Road Intersection Model and Traffic Model} \label{Subsec:RealisticTrafficModel}

%We study a more realistic road intersection as shown in Fig.~\ref{Fig:RoadIntersection}, and present our simulation results. 
We consider a more realistic road intersection model in Fig.~\ref{Fig:RoadIntersection}. There are totally 33 blocks with four entrances (block 26, 28, 30, and 32), and four exits (block 25, 27, 29, and 31). Vehicles in block $i \in \{2, 4, 6, 8\}$ go straight, turn left, turn right with the probabilities $g^s_i$, $g^l_i$, and $g^r_i$, such that $g^s_i + g^l_i + g^r_i = 1$. We assume vehicles can turn right when the traffic light is green.

Now, we describe the settings different from the last subsection. The discount factor $\gamma$ is $0.4 \sim 0.9$. The total UAV transmission power $P$ is set to $1 \sim 6$ W, which is used in \cite{fan2018UAV_Placement_Resource}. The total number of channels $C$ is 100 $\sim$ 200. It is much larger than that in Subsection \ref{Subsec:OptimalityVerification} since there are more vehicles in the realistic model. It is also commonly used such as \cite{hu2018uav_offloading}. The bandwidth of each channel $b$ is 5 KHz, therefore, the total bandwidth of all channels is $0.5 \sim 1$ MHz just like in \cite{zeng2018UAV_Trajectory}. The maximum power allocated to a vehicle $\rho_{\text{max}}$ is 0.9 W, and the maximum number of channels allocated to a vehicle $c_{\text{max}}$ is 50. The minimum and maximum height of the UAV is 10 meters and 200 meters. The probability of a vehicle going straight, turning left, and turning right ($g^s_i$, $g^l_i$, and $g^r_i$) is set to 0.4, 0.3, and 0.3, respectively, and each of them is assumed to be the same in block 2, 4, 6, and 8. We assume the arrival of vehicles in block 26, 28, 30, and 32 follows a binomial distribution with the same parameter $\lambda$ in the range $0.1 \sim 0.7$.

The UAV's horizontal and vertical flight actions are as follows. We assume that the UAV's block is 0 $\sim$ 8 since the number of vehicles in the intersection block 0 is generally the largest and the UAV will not move to the block far from the intersection block. Moreover, within a time slot we assume that the UAV can stay or only move to its adjacent blocks. The UAV's vertical flight action is set by (\ref{Eqn:UAVVerticalFlight}). In PowerControl, the UAV stays at block 0 with the height of 150 meters.

\subsection{5.3 Baseline Schemes}

We compare with two baseline schemes. Generally, the equal transmission power and channels allocation is common in communication systems for fairness. Therefore, they are used in baseline schemes.

The first baseline scheme is Cycle, i.e., the UAV cycles anticlockwise at a fixed height (e.g., 150 meters), and the UAV allocates the transmission power and channels equally to each vehicle in each time slot. The UAV moves along the fixed trajectory periodically, without considering the vehicle flows.

The second baseline scheme is Greedy, i.e., at a fixed height (e.g., 150 meters), the UAV greedily moves to the block with the largest number of vehicles. If a nonadjacent block has the largest number of vehicles, the UAV has to move to block 0 and then move to that block. The UAV also allocates the transmission power and the channels equally to each vehicle in each time slot. The UAV tries to serve the block with the largest number of vehicles by moving nearer to them.

There are several works which may be seen as comparisons. We take two works as examples. The first work is \cite{wu2018UAV_throughput}. Our work is different from \cite{wu2018UAV_throughput}, and the differences are as follows. 1) Our work uses RL methods since the variables are unknown. \cite{wu2018UAV_throughput} uses an optimization method based on all known variables. 2) In our work, all terminals move with their patterns. However, in \cite{wu2018UAV_throughput}, all terminals are stationary. 3) In our work, the UAV's action is 3D flight, and the UAV does not have fixed origin or destination. In \cite{wu2018UAV_throughput}, the the UAV's action is 2D flight with fixed origin and destination. Therefore, it does not need to compare them. 

The second work is \cite{fan2018UAV_Placement_Resource}. There is a fundamental difference between our work and \cite{fan2018UAV_Placement_Resource}. In \cite{fan2018UAV_Placement_Resource}, the channel states of UAV to all ground terminal links are LoS. However, in our work, the channel states change between LoS and NLoS links, which is a stochastic process. Therefore, it not suitable to compare them.

\subsection{5.4 Simulation Results}

The training time is about 4 hours, and the test time is almost real-time, since it only uses the well trained target policy network. Next, we first show the convergence of loss functions, and then show total throughput vs. discount factor, total transmission power, total number of channels and vehicle arrival probability, and finally present the total throughput and the UAV's flight time vs. energy percent for 3D flight.

\begin{figure}
  \centering
  \begin{minipage}{0.46\textwidth}
    \centering
    \includegraphics[height=0.73\linewidth,width=0.99\linewidth]{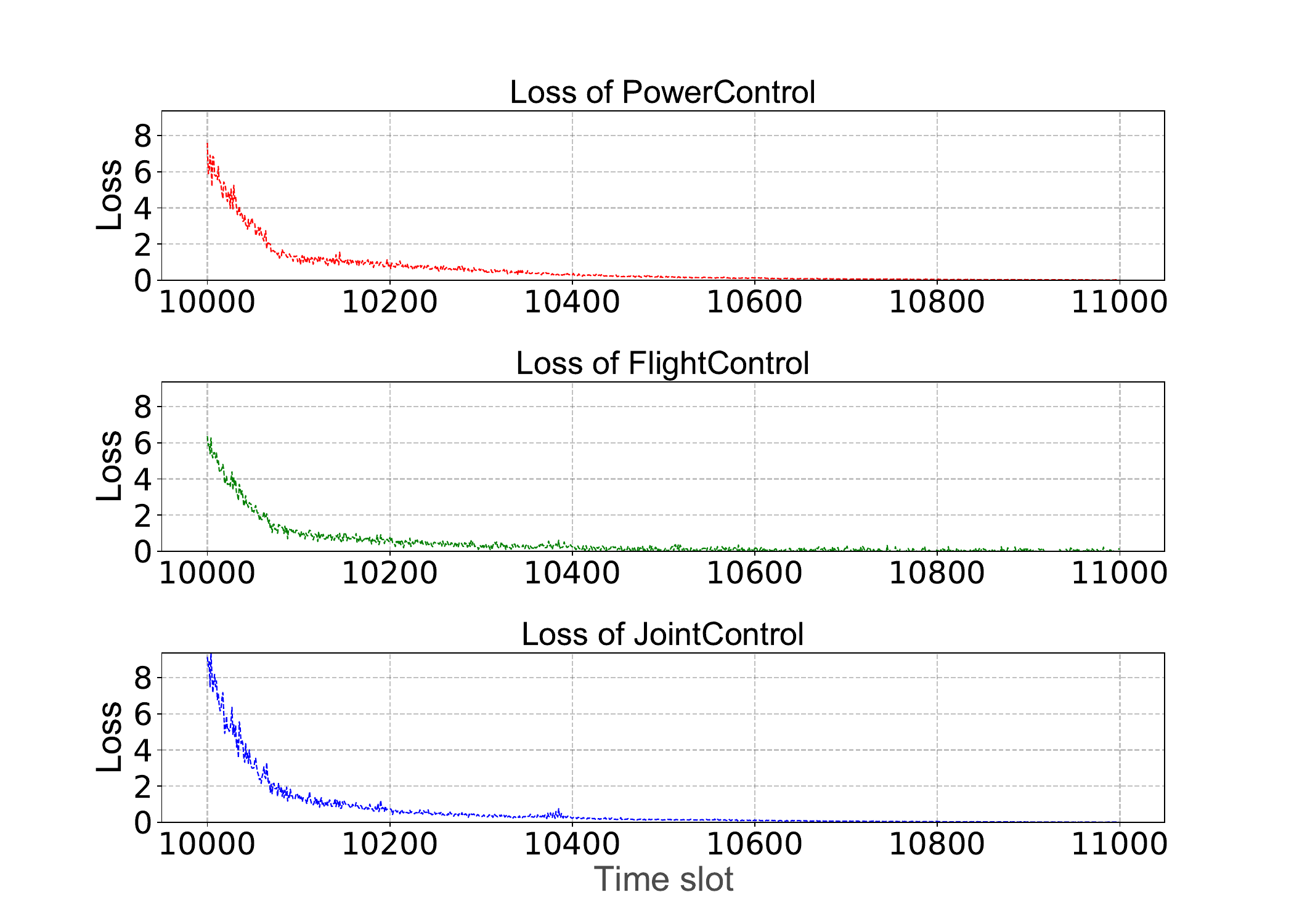}
    \caption{Convergence of loss functions in training stage.}
    \label{Fig:Loss}
  \end{minipage}%
  \hspace{0.2in}
  \begin{minipage}{0.46\textwidth}
    \centering
    \includegraphics[height=0.73\linewidth,width=0.99\linewidth]{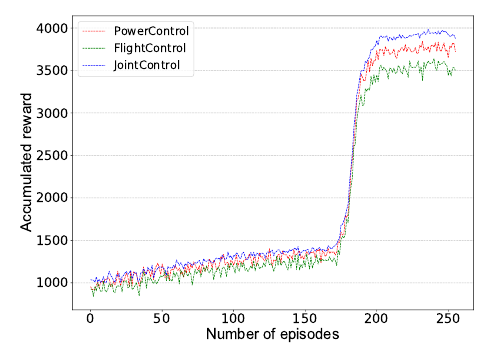}
    \caption{Cumulative reward vs. number of episodes.}
    \label{Fig:CumulativeReward}
  \end{minipage}%
\end{figure}

%\begin{figure}
%  \centering
%  \includegraphics[height=0.73\linewidth,width=0.99\linewidth]{Fig/CumulativeReward.pdf}
%  \caption{Cumulative reward vs. number of episodes.}
%  \label{Fig:CumulativeReward}
%\end{figure}

The convergence of loss functions in training stage for PowerControl, FlightControl, and JointControl indicates that the neural network is well-trained. It is shown in Fig.~\ref{Fig:Loss} when $P = 6$, $C = 200$, $\lambda = 0.5$ and $\gamma = 0.9$ during time slots 10,000 $\sim$ 11,000. The first 10,000 time slots are not shown since during the 0 $\sim$ 10,000, the experience replay buffer has not achieved its capacity. We see that, the loss functions in three algorithms converge after time slot 11,000. The other metrics in the paper are measured in test stage by default. The cumulative reward vs. number of episodes is shown in Fig.~\ref{Fig:CumulativeReward}. We see that, when the number of episodes approaches 256, the cumulative reward converges. This means that it is not needed to train the neural networks using more episodes since they have been well trained.

\begin{figure}
  \centering
  \begin{minipage}{0.32\textwidth}
    \centering
    \includegraphics[height=0.84\linewidth,width=0.93\linewidth]{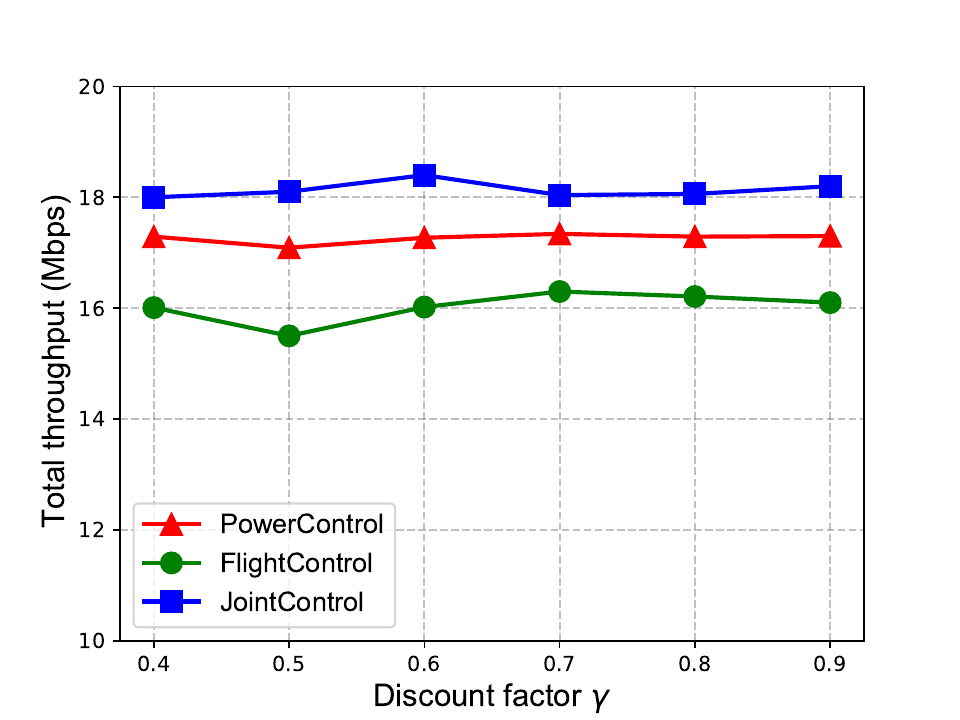}
    \caption{Throughput vs. discount factor $\gamma$.}
    \label{Fig:ThroughputWithDiscount}
  \end{minipage}%
  \hspace{0.05in}
  \begin{minipage}{0.33\textwidth}
    \centering
    \includegraphics[height=0.84\linewidth,width=0.93\linewidth]{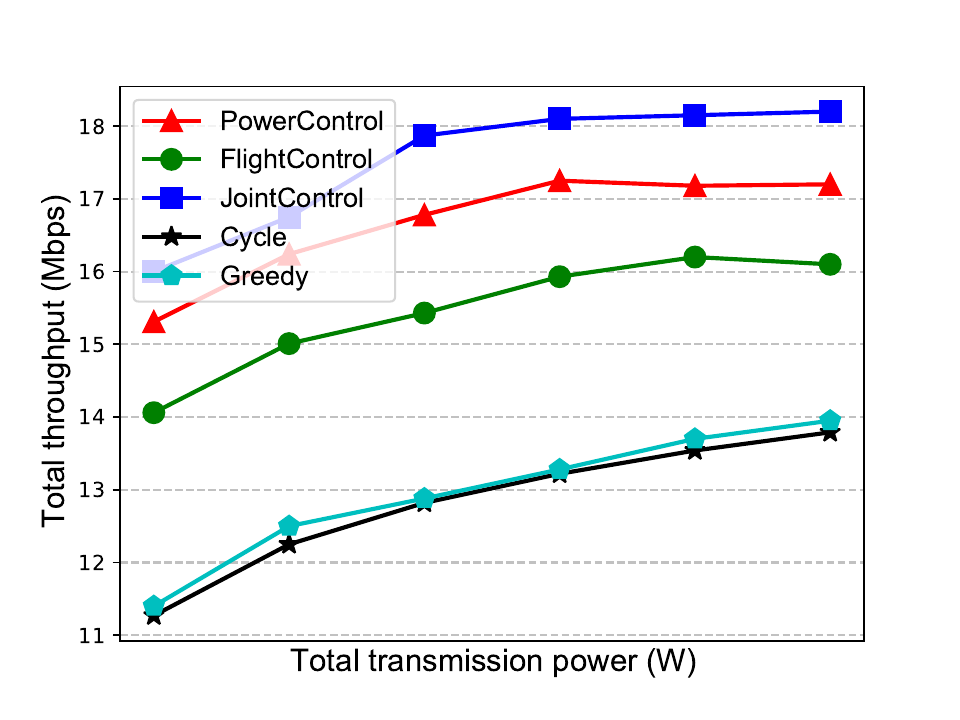}
    \caption{Total throughput vs. total transmission power (C = 200).}
    \label{Fig:ThroughputWithPower}
  \end{minipage}%
  \hspace{0.05in}
  \begin{minipage}{0.32\textwidth}
    \centering
    \includegraphics[height=0.84\linewidth,width=0.93\linewidth]{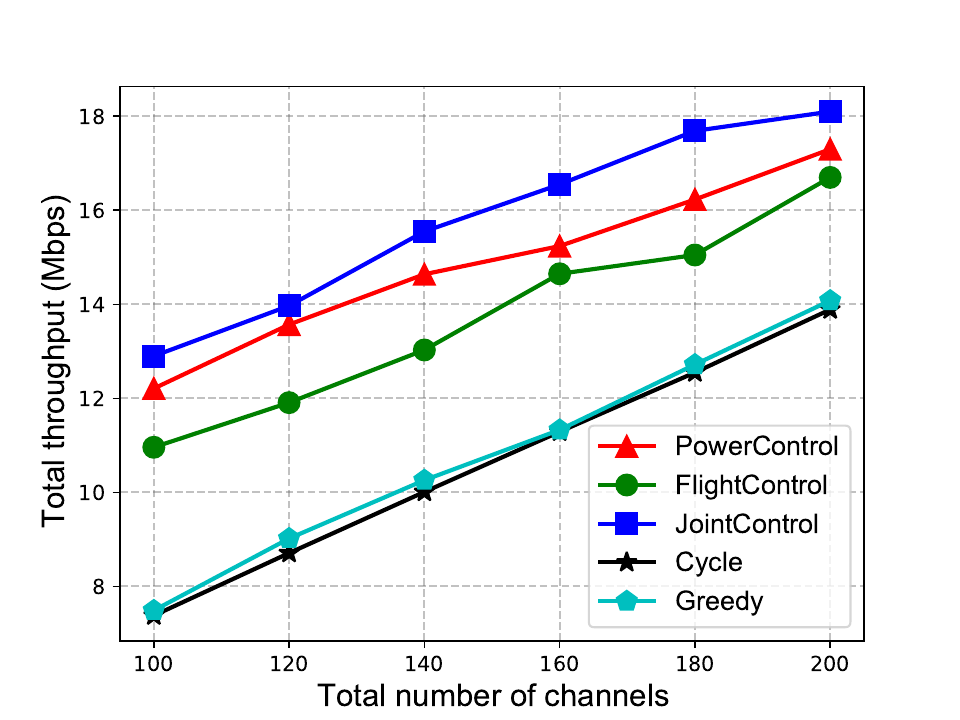}
    \caption{Total throughput vs. total number of channels (P = 6).}
    \label{Fig:ThroughputWithChannel}
  \end{minipage}%
\end{figure}

%\begin{figure}
%  \centering
%  \includegraphics[height=0.84\linewidth,width=0.93\linewidth]{Fig/ThroughputWithDiscount.pdf}
%  \caption{Throughput vs. discount factor $\gamma$.}
%  \label{Fig:ThroughputWithDiscount}
%\end{figure}

Total throughput vs. discount factor $\gamma$ is drawn in Fig.~\ref{Fig:ThroughputWithDiscount} when $P = 6$, $C = 200$, and $\lambda = 0.5$. We can see that, when $\gamma$ changes, the throughput of three algorithms is steady. The discount factor does not noticeably affect the performances of the algorithms. This implies that the discount factor does not play an important role in the training process for the neural networks. If the neural networks are well trained, the proposed algorithms will get steady. JointControl achieves higher total throughput, compared with PowerControl and FlightControl, respectively. PowerControl achieves higher throughput than FlightControl since PowerControl allocates power and channel to strongest channels while FlightControl only adjusts the UAV's 3D position to enhance the strongest channel and the equal power and channel allocation is far from the best strategy in OFDMA.

%\begin{figure}
%  \centering
%  \includegraphics[height=0.84\linewidth,width=0.93\linewidth]{Fig/ThroughputWithPower.pdf}
%  \caption{Total throughput vs. total transmission power (C = 200).}
%  \label{Fig:ThroughputWithPower}
%\end{figure}

%\begin{figure}
%  \centering
%  \includegraphics[height=0.84\linewidth,width=0.93\linewidth]{Fig/ThroughputWithChannel.pdf}
%  \caption{Total throughput vs. total number of channels (P = 6).}
%  \label{Fig:ThroughputWithChannel}
%\end{figure}

Total throughput vs. total transmission power ($P = 1 \sim 6$) and total number of channels ($C = 100 \sim 200$) are shown by Fig.~\ref{Fig:ThroughputWithPower} and Fig.~\ref{Fig:ThroughputWithChannel}, where we set $\lambda = 0.5$ and $\gamma = 0.9$. We see that JointControl achieves the best performance for different transmission power and channel budgets, respectively. Moreover, the total throughput of all algorithms increases when the total transmission power or total number of channels increases. PowerControl and FlightControl only adjust the transmission power or 3D flight, while JointControl jointly adjusts both of them, so its performance is the best. The total throughput of DDPG-based algorithms is improved greatly than that of Cycle and Greedy. The performance of Greedy is a little better than Cycle, since Greedy tries to get nearer to the block with the largest number of vehicles.

\begin{figure}
  \centering
  \begin{minipage}{0.32\textwidth}
    \centering
    \includegraphics[height=0.84\linewidth,width=0.93\linewidth]{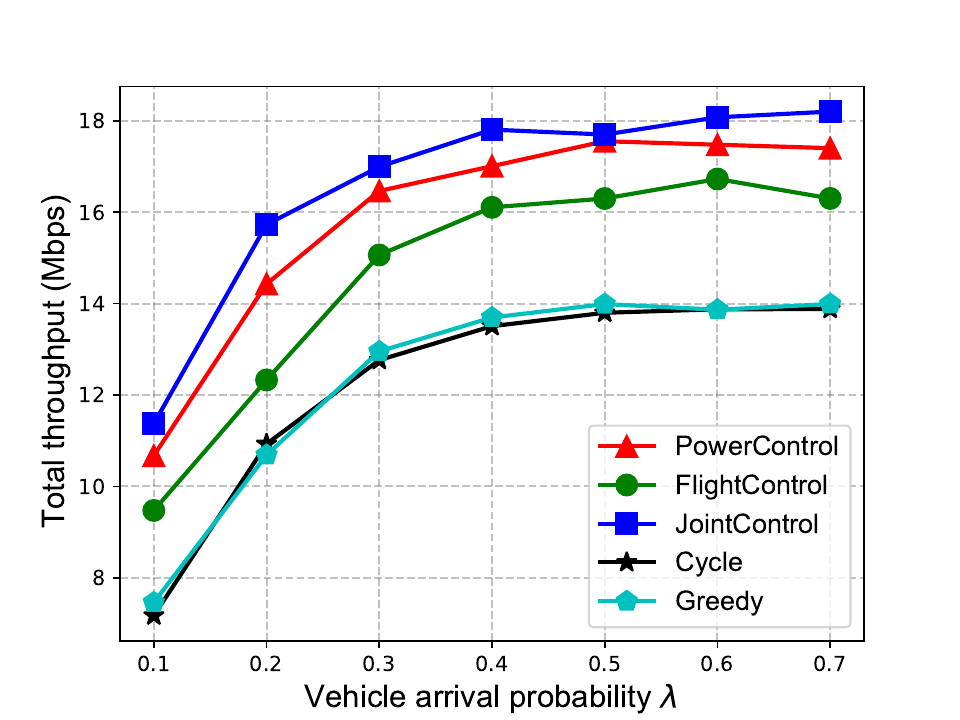}
    \caption{Total throughput vs. vehicle arrival probability $\lambda$.}
    \label{Fig:ThroughputWithLambda}
  \end{minipage}%
  \hspace{0.05in}
  \begin{minipage}{0.32\textwidth}
    \centering
    \includegraphics[height=0.84\linewidth,width=0.93\linewidth]{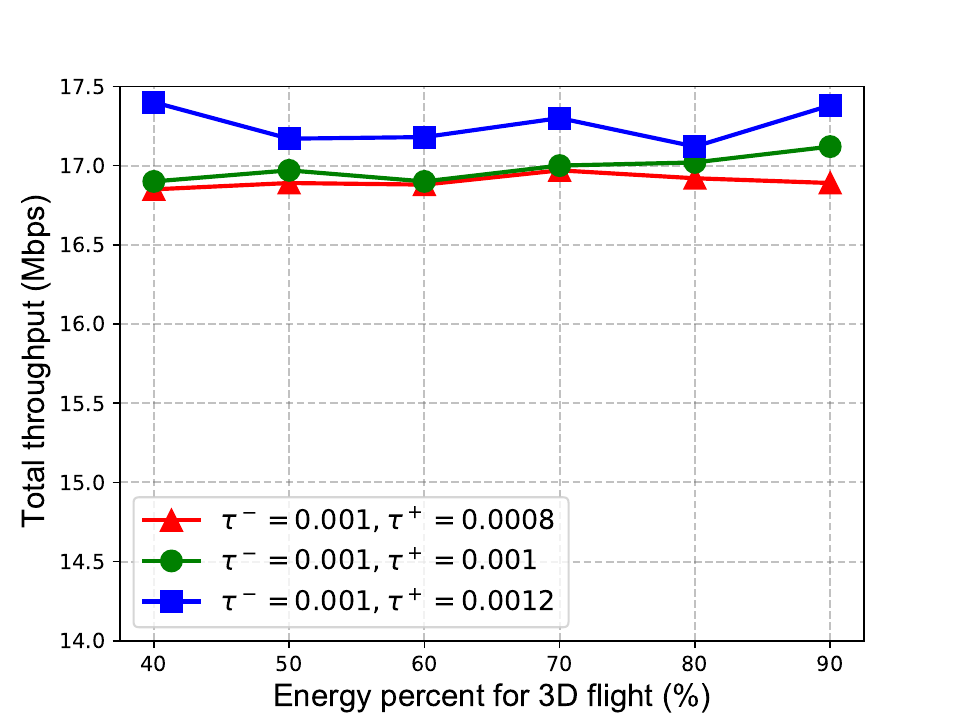}
    \caption{Total throughput vs. energy percent for 3D flight in JointControl (P = 6, C = 200).}
    \label{Fig:ThroughputWithEnergyPercent}
  \end{minipage}%
  \hspace{0.05in}
  \begin{minipage}{0.32\textwidth}
    \centering
    \includegraphics[height=0.84\linewidth,width=0.93\linewidth]{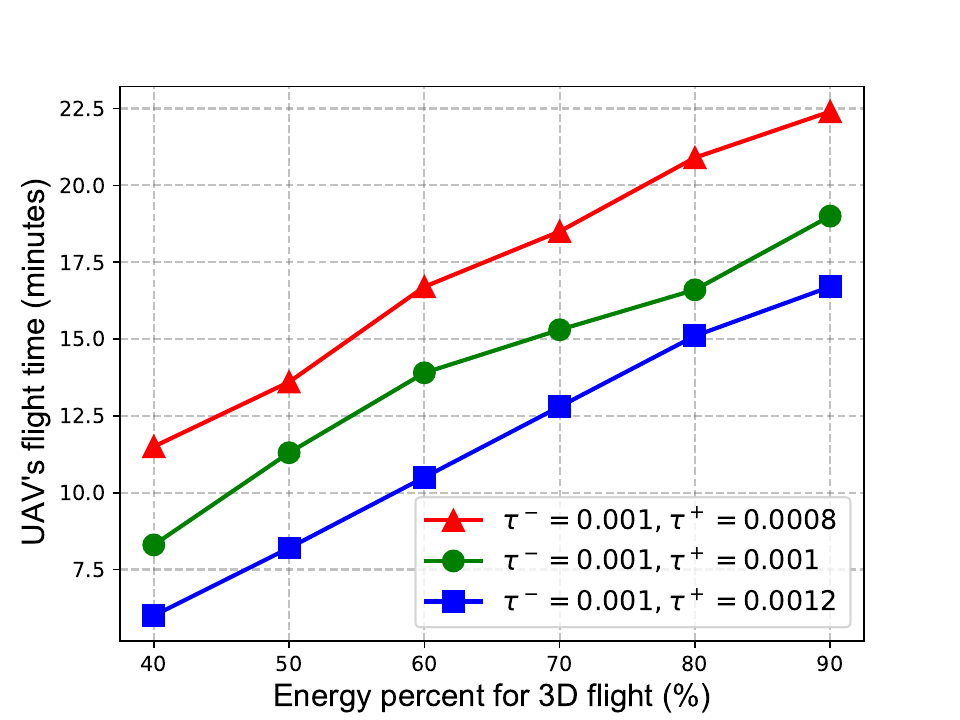}
    \caption{UAV's flight time vs. energy percent for 3D flight in JointControl (P = 6, C = 200).}
    \label{Fig:FlightTimeWithEnergyPercent}
  \end{minipage}%
\end{figure}

%\begin{figure}
%  \centering
%  \includegraphics[height=0.84\linewidth,width=0.93\linewidth]{Fig/ThroughputWithLambda.pdf}
%  \caption{Total throughput vs. vehicle arrival probability $\lambda$.}
%  \label{Fig:ThroughputWithLambda}
%\end{figure}

Total throughput vs. vehicle arrival probability $\lambda$ is shown in Fig.~\ref{Fig:ThroughputWithLambda}. Note that the road intersection has a capacity of $2$ units, i.e., it can serve at most two traffic flows at the same time, therefore, it cannot serve traffic flows where $\lambda$ is very high, e.g., $\lambda = 0.8$ and $\lambda = 0.9$. We see that, when $\lambda$ increases, i.e., more vehicles arrive at the intersection, the total throughput increases. However, when $\lambda$ gets higher, e.g., $\lambda = 0.6$, the total throughput saturates due to traffic congestion.

%\begin{figure}
%  \centering
%  \includegraphics[height=0.84\linewidth,width=0.93\linewidth]{Fig/ThroughputWithEnergyPercent.pdf}
%  \caption{Total throughput vs. energy percent for 3D flight in JointControl (P = 6, C = 200).}
%  \label{Fig:ThroughputWithEnergyPercent}
%\end{figure}

%\begin{figure}
%  \centering
%  \includegraphics[height=0.84\linewidth,width=0.93\linewidth]{Fig/FlightTimeWithEnergyPercent.pdf}
%  \caption{UAV's flight time vs. energy percent for 3D flight in JointControl (P = 6, C = 200).}
%  \label{Fig:FlightTimeWithEnergyPercent}
%\end{figure}

Next, we test the metrics considering of the energy consumption of 3D flight. The total throughput vs. energy percent for 3D flight in JointControl is shown in Fig.~\ref{Fig:ThroughputWithEnergyPercent}. When $\tau^+$ increases, the total throughput almost increases and has more variance since the UAV prefers to get higher throughput through more movements. We can also see a tradeoff of energy consumption and throughput in Fig.~\ref{Fig:ThroughputWithEnergyPercent}. The UAV's total energy consumption is separated to two parts, transmission and 3D flight. The UAV's total transmission power among all vehicles under different learning rates is almost the same. If $\tau^+$ increases, the UAV moves more; therefore, the energy consumption increases. We see that, when $\tau^+$ = 0.0012, the throughput is much higher than the other two cases, i.e., $\tau^+$ = 0.001 and 0.0008. When $\tau^+$ = 0.001, the throughput is almost the same as the case $\tau^+$ = 0.0008. It implies that, when $\tau^+$ is high, such as 0.0012, the throughput will be much higher than the other cases since the UAV can improve the total throughput by more energy consumption (or movement); however, when $\tau^+$ is no larger than 0.001, the throughput almost stays the same, since more energy consumption (or movement) almost cannot improve the total throughput.

The UAV's flight time vs. energy percent for 3D flight in JointControl is shown in Fig.~\ref{Fig:FlightTimeWithEnergyPercent}. When $\tau^- = 0.001$ and $\tau^+ = 0.0008$, the UAV's flight time is the longest since the UAV is inactive. When $\tau^- = 0.001$ and $\tau^+ = 0.0012$, the UAV's flight time is the shortest, since the UAV is active and prefers to flight. When $\tau^- = \tau^+ = 0.001$, the UAV's flight time is between the other two cases. If the energy percent for 3D flight increases, the UAV's flight time increases linearly in the three cases.

We can see that proposed algorithms and comparisons have the same shape in several figures (e.g., Fig.~\ref{Fig:ThroughputWithPower}, Fig.~\ref{Fig:ThroughputWithChannel}, and Fig.~\ref{Fig:FlightTimeWithEnergyPercent}). This is because the neural networks are well trained. From the results of Cycle and Greedy, we also conclude that, if the UAV's height is fixed, the performance of Cycle and Greedy is almost the same. This is because the UAV almost cannot improve the communication performance if its movement and communication control policies are fixed.

\section{6 Conclusions} \label{Sec:Conclusion}

We studied a UAV-assisted vehicular network where the UAV acted as a relay to maximize the total throughput between the UAV and vehicles. We focused on the downlink communication where the UAV could adjust its transmission control (power and channel) under 3D flight. We formulated our problem as a MDP problem, explored the state transitions of UAV and vehicles under different actions, and then proposed three DDPG based algorithms, and finally extended them to account for the energy consumption of the UAV's 3D flight by modifying the reward function and the DDPG framework. In a simplified scenario with small state space and action space, we verified the optimality of DDPG-based algorithms. Through simulation results, we demonstrated the superior performance of the algorithms under a more realistic traffic scenario compared with two baseline schemes.

In the future, we will consider the scenario where multiple UAVs constitute a relay network to assist vehicular networks and study the coverage overlap/probability, relay selection, energy harvesting communications, and UAV cooperative communication protocols. We pre-trained the proposed solutions using servers, and we hope the UAV trains the neural netwroks in the future if light and low energy consumption GPUs are applied at the edge.

    %\newpage
    \section{Acknowledgement}
    Ming~Zhu was supported by National Natural Science Foundations of China (Grant No. 61902387).

%    \bibliographystyle{ACM-Reference-Format}
%    \bibliography{Bib.bib}


%    \input{7Supplement.tex}
\clearpage
\centerline{\large{\textbf{Supplementary material}}}

% \begin{eqnarray*}
% && \textbf{Supplement} \nonumber
% \end{eqnarray*}
% \begin{center}
% \begin{equation*}
% \textbf{Supplement} \nonumber
% \end{equation*}
% \end{center}
% \begin{center}
%   \large{\textbf{Supplement}}
% \end{center}

% \title{Supplement}

\vspace{0.05in}
\begin{table*}
\normalsize
\centering
\begin{tabular}[h]{lp{0.46\textwidth}}
  \toprule
  \textbf{Algorithm 3}: Q-learning-based algorithm \\
  \hline
  \,\,\,~\textbf{Input}: the number of episodes $K$, the learning rate $\alpha$, parameter $\epsilon$.\\
  \,\,\,1:~Initialize all states. Initialize $Q(s, a)$ for all state-action pairs randomly.\\
  \,\,\,2:~\textbf{for} episode $k = 1$ to $K$ \\
  \,\,\,3:~~~Observe the initial state $s_1$. \\
  \,\,\,4:~~~\textbf{for} each slot $t = 1$ to $T$ \\
  \,\,\,5:~~~~~Select the UAV's action $a_t$ from state $s_t$ using (\ref{Eqn:EpsilonGreedy}). \\
  \,\,\,6:~~~~~Execute the UAV's action $a_t$, receive reward $r_t$, and observe a new state $s_{t+1}$ from the environment.\\
  \,\,\,7:~~~~~Update Q-value function: $Q(s_t, a_t) \leftarrow Q(s_t, a_t) + \alpha \left[ r_t + \gamma \max_{a_{t + 1}} Q(s_{t + 1}, a_{t + 1}) - Q(s_t, a_t) \right]$. \\
  \bottomrule
\end{tabular}
\end{table*}

\begin{figure*}[h]
  \centering
  \includegraphics[height=0.55\linewidth,width=0.47\linewidth]{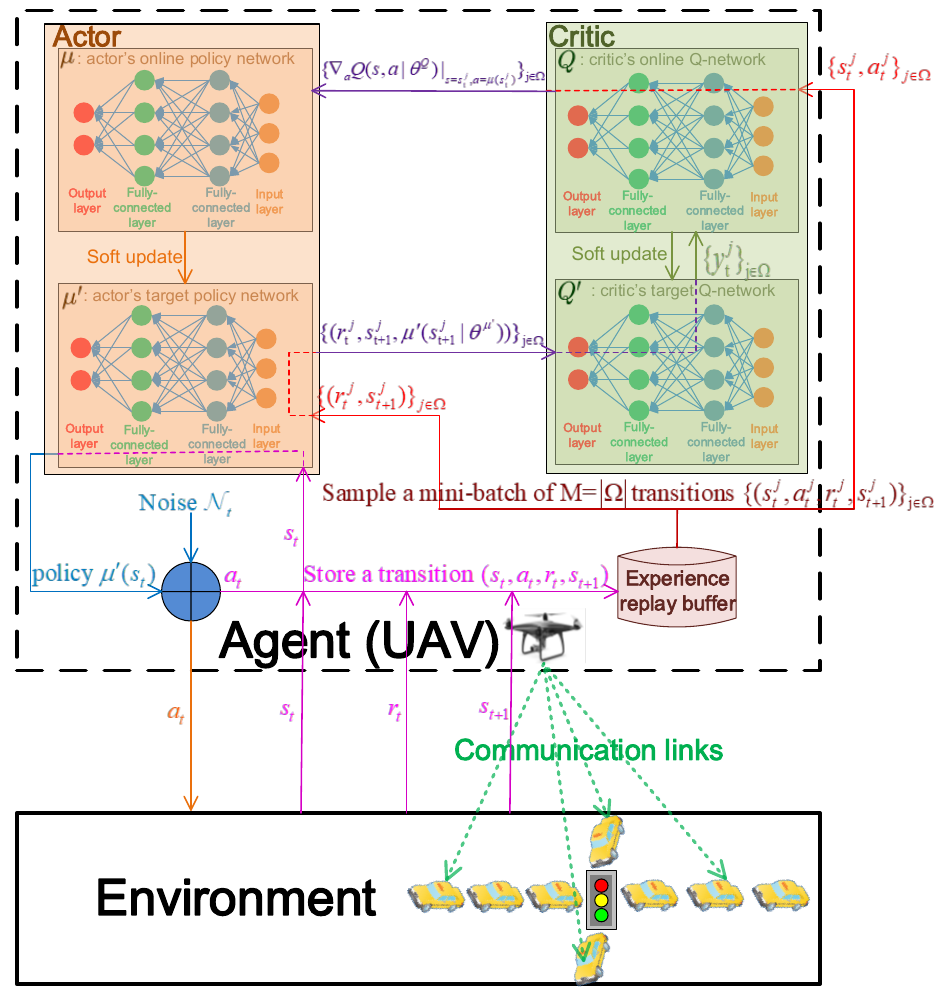}
  \caption{Framework of the DDPG algorithm.}
  \label{Fig:DDPG_Framework}
\end{figure*}

% \vspace{0.3in}

% \clearpage

\noindent{\textbf{1)~Q-learning}}

The state transition probabilities of MDP are unknown in our problem, since some variables are unknown, e.g., $\alpha_1$, $\alpha_2$, $\lambda_1$, and $\lambda_2$. Our problem cannot be solved directly using conventional MDP solutions, e.g., dynamic programming algorithms, policy iteration and value iteration algorithms. Therefore, we apply the deep reinforcement learning approach. The return from a state is defined as the sum of discounted future reward $\sum^{T}_{i = t} \gamma^{i - t} r(s_i, a_i)$, where $T$ is the total number of time slots, and $\gamma \in (0, 1)$ is a discount factor that diminishes the future reward and ensures that the sum of an infinite number of rewards is still finite. Let $Q^{\pi}(s_t, a_t) = \mathbb{E}_{a_i \sim \pi}[\sum^{T}_{i = t} \gamma^{i - t} r(s_i, a_i) |s_t, a_t]$ represents the expected return after taking action $a_t$ in state $s_t$ under policy $\pi$. The Bellman equation gives the optimality condition in conventional MDP solutions \cite{sutton2018RL}:
\begin{eqnarray}
Q^{\pi}(s_t, a_t) \! = \! \! \! \sum_{s_{t \! + \! 1}, r_t} p(s_{t \! + \! 1}, r_t|s_t, a_t)  \! \left[ r_t \! + \! \gamma \max_{a_{t \! + \! 1}} Q^{\pi} (s_{t \! + \! 1}, a_{t \! + \! 1}) \right]. \nonumber
\end{eqnarray}

Q-learning \cite{watkins1992Q_learning} is a classical model-free RL algorithm \cite{Wirth2016ModelFree}. Q-learning with the essence of exploration and exploitation aims to maximize the expected return by interacting with the environment. The update of $Q(s_t, a_t)$ is
\begin{eqnarray}
&& \hspace{-0.01in} Q(s_t,  a_t)  \leftarrow  Q(s_t,  a_t)  +  \alpha  [ r_t  + \gamma \max_{a_{t + 1}} Q(s_{t  +  1}, a_{t + 1})  \nonumber \\
&& \hspace{1.51in} -  Q(s_t, a_t) ], \label{Eqn:UpdateQInQLearning}
\end{eqnarray}
where $\alpha$ is a learning rate.

Q-learning uses the $\epsilon$-greedy strategy \cite{van2016DoubleQLearning} to select an action, so that the agent behaves greedily most of the time, but selects randomly among all the actions with a small probability $\epsilon$. The $\epsilon$-greedy strategy is defined as follows
\begin{eqnarray}
&& \hspace{-0.4in} a_t = \label{Eqn:EpsilonGreedy}
\begin{cases}
\arg\max_a Q(s_t, a) ,  ~\text{with probability}~1 - \epsilon,  \\
\text{a random action},  ~~\, ~\text{with probability}~\epsilon.
\end{cases}
\end{eqnarray}

The Q-learning algorithm \cite{sutton2018RL} is shown in Alg. 3. Line 1 is initialization. In each episode, the inner loop is executed in lines 4 $\sim$ 7. Line 5 selects an action using (\ref{Eqn:EpsilonGreedy}), and then the action is executed in line 6. Line 7 updates the Q-value.

\newpage
\noindent{\textbf{2)~Framework of the DDPG algorithm}}

The framework of the DDPG algorithm is shown in Fig.~\ref{Fig:DDPG_Framework}. 

\textbf{1)} In the training stage, we train the actor and the critic, and store the parameters of their neural networks. The training stage has two parts. First, $Q$ and $\mu$ are trained through a random mini-batch of transitions sampled from the experience replay buffer $R_b$. Secondly, $Q'$ and $\mu'$ are trained through soft update.

The training process is as follows. A mini-batch of $M$ transitions $\{ (s^j_t, a^j_t, r^j_t, s^j_{t + 1}) \}_{j \in \Omega}$ are sampled from $R_b$, where $\Omega$ is a set of indices of sampled transitions from $R_b$ with $|\Omega| = M$. Then two data flows are outputted from $R_b$: $\{ r^j_t, s^j_{t + 1} \}_{j \in \Omega} \rightarrow \mu'$, and $\{ s^j_t, a^j_t \}_{j \in \Omega} \rightarrow Q$. $\mu'$ outputs $\{ r^j_t, s^j_{t + 1}, \mu'(s^j_{t + 1}|\theta^{\mu'}) \}_{j \in \Omega}$ to $Q'$ to calculate $\{ y^j_t \}_{j \in \Omega}$. Then $Q$ calculates and outputs $\{ \nabla_{a} Q(s, a| \theta^Q)|_{s = s^j_t, a = \mu(s^j_t)} \}_{j \in \Omega}$ to $\mu$. $\mu$ updates its parameters by (\ref{Eqn:NablaQ}). Then two soft updates are executed for $Q'$ and $\mu'$ in (\ref{Eqn:QUpdate}) and (\ref{Eqn:MuUpdate}), respectively.

The data flow of the critic's target Q-network $Q'$ and online Q-network $Q$ are as follows. $Q'$ takes $\{ (r^j_t, s^j_{t + 1}, \mu'(s^j_{t + 1}|\theta^{\mu'})) \}_{j \in \Omega}$ as the input and outputs $\{ y^j_t \}_{j \in \Omega}$ to $Q$. $y^j_t$ is calculated by \eqref{Eqn:y}.

$Q$ takes $\left\{ \{ s^j_t, a^j_t \}_{j \in \Omega}, \right \}$ as the input and outputs $\{ \nabla_{a} Q(s, a| \theta^Q)|_{s = s^j_t, a = \mu(s^j_t)} \}_{j \in \Omega}$ to $\mu$ for updating parameters in (\ref{Eqn:NablaQ}), where $\{ s^j_t \}_{j \in \Omega}$ are sampled from $R_b$, and $\mu(s^j_t) = \arg\max_a Q(s^j_t, a)$.

The data flows of the actor's online policy network $\mu$ and target policy network $\mu'$ are as follows. After $Q$ outputs $\{ \nabla_{a} Q(s, a| \theta^Q)|_{s = s^j_t, a = \mu(s^j_t)} \}_{j \in \Omega}$ to $\mu$, $\mu$ updates its parameters by (\ref{Eqn:NablaQ}). $\mu'$ takes $\{ r^j_t, s^j_{t + 1} \}_{j \in \Omega}$ as the input and outputs $\{ r^j_t, s^j_{t + 1}, \mu'(s^j_{t + 1}|\theta^{\mu'}) \}_{j \in \Omega}$ to $Q'$ for calculating $\{ y^j_t \}_{j \in \Omega}$ in (\ref{Eqn:y}), where $\{ r^j_t, s^j_{t + 1} \}_{j \in \Omega}$ are sampled from $R_b$.

The updates of parameters of four neural networks ($Q$, $Q'$, $\mu$, and $\mu'$) are as follows. The online Q-network $Q$ updates its parameters by minimizing the $L_2$-norm loss function $\text{Loss}_t(\theta^Q)$ to make its Q-value fit $y^j_t$.

The target Q-network $Q'$ updates its parameters $\theta^{Q'}$ by (\ref{Eqn:QUpdate}). The online policy network $\mu$ updates its parameters following \eqref{Eqn:NablaQ}. The target policy network $\mu'$ updates its parameters $\theta^{\mu'}$ by (\ref{Eqn:MuUpdate}).

In each time slot $t$, the current state $s_t$ from the environment is delivered to $\mu'$, and $\mu'$ calculates the UAV's target policy $\mu'(s_t)|\theta^{\mu'}$. Finally, an exploration noise $\mathcal{N}$ is added to $\mu'(s_t|\theta^{\mu'})$ to get the UAV's action in (\ref{Eqn:NewAction}).

\textbf{2)} In the test stage, we restore the neural network of the actor's target policy network $\mu'$ based on the stored parameters. This way, there is no need to store transitions to the experience replay buffer $R_b$. Given the current state $s_t$, we use $\mu'$ to obtain the UAV's optimal action $\mu'(s_t|\theta^{\mu'})$. Note that there is no noise added to $\mu'(s_t|\theta^{\mu'})$, since all neural networks have been trained and the UAV has got the optimal action through $\mu'$. Finally, the UAV executes the action $\mu'(s_t|\theta^{\mu'})$.

\end{document}